\documentclass{article}

 \usepackage[main, final]{neurips_2026}
\usepackage[utf8]{inputenc} % allow utf-8 input
\usepackage[T1]{fontenc}    % use 8-bit T1 fonts
\usepackage{hyperref}       % hyperlinks
\usepackage{url}            % simple URL typesetting
\usepackage{booktabs}       % professional-quality tables
\usepackage{amsfonts}       % blackboard math symbols
\usepackage{nicefrac}       % compact symbols for 1/2, etc.
\usepackage{microtype}      % microtypography
\usepackage{xcolor}         % colors
\usepackage{graphicx}
\usepackage{amsmath}
\usepackage{amssymb}
\usepackage{mathtools}
\usepackage{multirow}
\usepackage{cleveref}
\usepackage{wrapfig}
\usepackage{bbm}
\usepackage{subcaption}
\title{CytoWave: Perturbation-Centric Pretraining for Single-Cell Response Prediction}

\author{%
  \textbf{Yuwei Miao}\textsuperscript{1,*},
  \textbf{Jingquan Yan}\textsuperscript{1,*},
  \textbf{Hehuan Ma}\textsuperscript{1},
  \textbf{Thao M. Dang}\textsuperscript{1}, \\
  \textbf{Lin Xu}\textsuperscript{2},
  \textbf{Siyuan Zhang}\textsuperscript{2},
  \textbf{Junzhou Huang}\textsuperscript{1} \\
  \\
  \textsuperscript{1}Department of Computer Science and Engineering, \\
  University of Texas at Arlington, Arlington, TX, USA \\
  \textsuperscript{2}UT Southwestern Medical Center, Dallas, TX, USA \\
  \texttt{\{yuwei.miao,jingquan.yan,hehuan.ma,thaomai.dang,jzhuang\}@uta.edu} \\
  \texttt{\{lin.xu,siyuan.zhang\}@utsouthwestern.edu} \\
  \textsuperscript{*}Equal contribution.
}

\begin{document}

\maketitle

\begin{abstract}
Predicting gene expression under perturbations from single-cell data is a fundamental task for modeling cellular responses, with applications in discovering gene function and designing therapeutics. However, existing methods often struggle to generalize across datasets due to differences in experimental protocols, biological contexts, and species. We present \textbf{CytoWave}, a perturbation-centric pretraining framework for single-cell genetic perturbation response prediction. We leverage large-scale cross-species perturbation datasets unified via orthology and train CytoWave in a two-stage manner. In the first stage, we learn reference control cell states via masked gene expression recovery on unperturbed cells. In the second stage, we pretrain the model to predict perturbed gene expression given control states and perturbation signals, together with a contrastive objective that aligns latent representations of perturbation responses with perturbation signals. We perform strict dataset-held-out evaluation across multiple condition combinations and case studies of pathway activity changes, demonstrating consistent and effective performance of CytoWave in predicting single-cell genetic perturbation responses.
\end{abstract}

\section{Introduction}
Single-cell genetic perturbation experiments measure transcriptomic responses to targeted gene perturbations at single-cell resolution, providing a rich data source for characterizing gene function, regulatory relationships, and cellular state variation. Compared to unperturbed single-cell atlases, perturbation data links perturbation targets to expression responses, enabling comparison of perturbation responses across cellular contexts and experimental conditions~\cite{dixit2016perturb, kampmann2018crispri}. These perturbation–response data provide a rich basis for learning generalizable representations of gene function and cellular response patterns~\cite{lotfollahi2019scgen, lotfollahi2021compositional}. Advances in Perturb-seq and CRISPR-based technologies have led to the accumulation of large-scale perturbation datasets with increasing gene coverage and experimental diversity, supporting cross-dataset and cross-condition pretraining~\cite{dixit2016perturb, kampmann2018crispri, Zhu2025.12.23.696273, peidli2024scperturb}. In this context, a variety of methods have been developed for modeling single-cell perturbation responses.

Existing methods for single-cell perturbation modeling have demonstrated strong potential in capturing gene expression responses and highlight the importance of accurately predicting cellular responses after genetic perturbations~\cite{lotfollahi2019scgen, cui2024scgpt, lotfollahi2023predicting, roohani2024predicting, chen2024genepert, hao2024large}. These approaches are often developed for specific datasets or experimental conditions using supervised training, or applied from general single-cell models to perturbation tasks through downstream fine-tuning. Recent CRISPR-based wet-lab experiments have covered a broader set of perturbed genes, achieved deeper read depth, and incorporated diverse experimental conditions, contributing to the increasing scale and diversity of perturbation data~\cite{papalexi2021characterizing, tian2021genome, adamson2016multiplexed, datlinger2021ultra, mcfarland2020multiplexed, peidli2024scperturb, Zhu2025.12.23.696273}. This evolving data landscape makes it possible and promising to extend existing approaches toward perturbation-centric pretraining, motivating the unification of large-scale datasets across diverse experimental conditions, gene identifiers, and species.

With this growing data landscape, an important question is how to represent perturbation responses in a way that supports generalization across diverse cellular contexts. Modeling single-cell genetic perturbations requires going beyond fitting individual gene expression responses. Biologically, the transcriptional state of a cell is governed by gene regulatory programs that shape expression patterns across functionally related genes~\cite{lopez2018deep, luecken2022benchmarking}. Perturbations propagate through these programs to induce global transcriptional responses rather than isolated gene-level changes~\cite{dixit2016perturb, adamson2016multiplexed}. Regulatory programs are often shared across cellular contexts, and perturbations acting through similar regulatory mechanisms can induce related patterns of cellular response across different cells. These observations suggest that perturbation responses depend on both the initial cellular state and the applied perturbation. Rather than modeling this relationship independently for each cell, we parameterize perturbation responses within a shared latent representation space, where perturbations are represented through cellular-state-dependent response patterns. This provides a unified way to represent related perturbation response patterns across different cellular contexts.

To address these challenges, we propose \textbf{CytoWave}, a perturbation-centric pretraining framework for single-cell genetic perturbation modeling. We begin by systematically collecting and organizing large-scale perturbation datasets across studies and species, and unify them through orthology-based gene alignment and standardized metadata vocabularies. This enables pretraining on diverse perturbation data under a unified setting, with evaluation datasets held out from pretraining. 
Building on this unified data foundation, CytoWave models perturbation-dependent changes in latent cellular states.
The model is trained in a two-stage manner. First, it learns reference control cell representations from unperturbed data. 
It then predicts perturbed gene expression from control states and perturbation signals, together with a contrastive objective that aligns latent state changes with perturbation signals. 
Together, these design choices define a unified framework for learning perturbation-aware representations and support generalization across genes and experimental conditions. 
The main contributions of this work can be summarized as follows:
\begin{itemize}
\item We collect and unify 42 single-cell genetic perturbation datasets across 19 studies and 2 species using orthology-based gene alignment and standardized metadata, resulting in a corpus of 24.7 million cells for pretraining and dataset-held-out benchmarking.
\item We introduce CytoWave, a perturbation-centric pretraining framework learning perturbation-aware representations by modeling perturbation-dependent changes in latent cellular states.
\item CytoWave generalizes under dataset-held-out evaluation, accurately predicting genetic perturbation responses and capturing perturbation-associated pathway activity changes.
\end{itemize}

\begin{figure*}[t!]
  \centering
  \includegraphics[width=\textwidth]{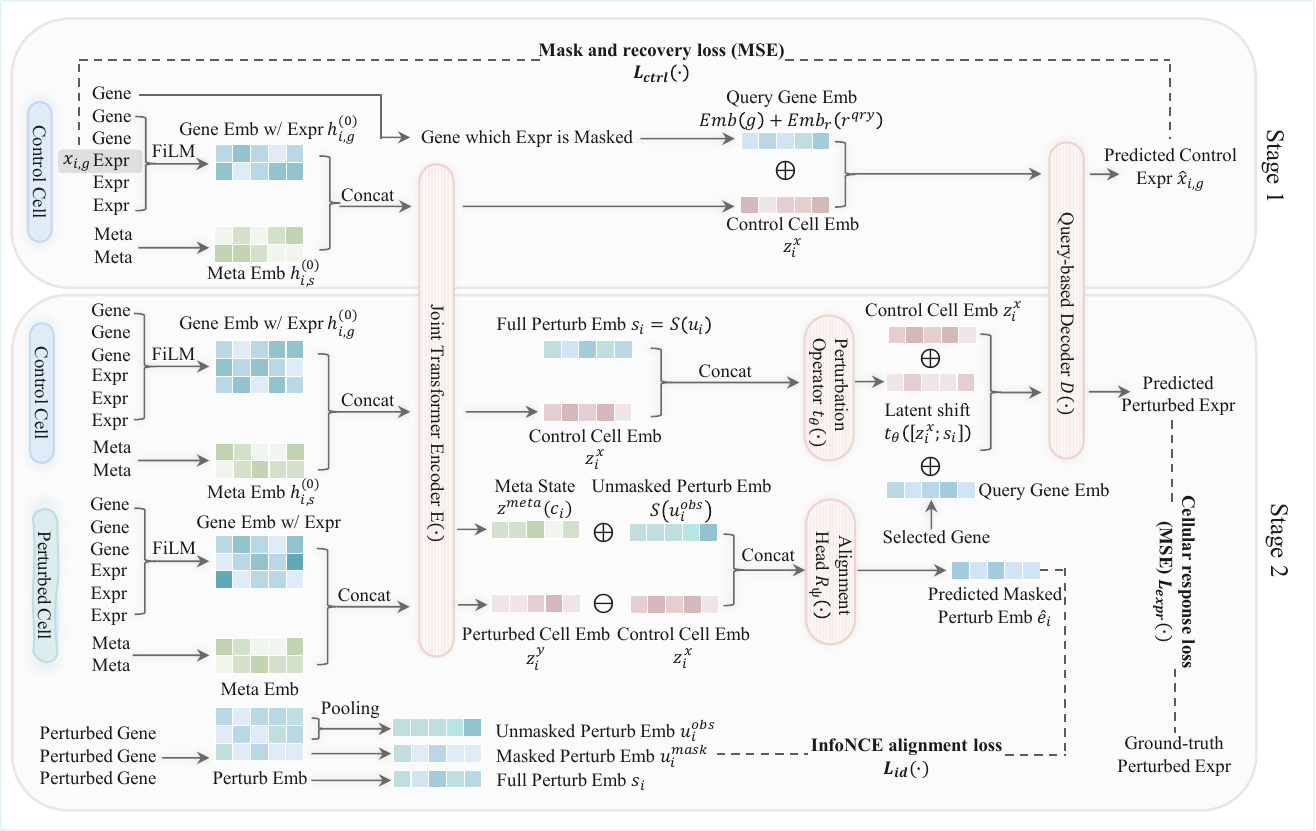}
  \caption{CytoWave architecture and two-stage self-supervised pretraining. \textbf{Stage 1}: Learning control representations via masked gene expression and recovery using a joint Transformer encoder with FiLM-conditioned gene embeddings. \textbf{Stage 2}: Learning a genetic perturbation operator by jointly predicting cellular responses after perturbation (MSE loss) and contrastive perturbation alignment, where predicted perturbation embeddings are aligned with ground-truth masked perturbation embeddings with in-batch InfoNCE loss.}
  \label{fig:overall_pipeline}
\vspace{-15pt}
\end{figure*}

\section{Related Work}

\subsection{Single-cell Pretrained Models in Perturbation}
In recent years, several single-cell pretrained models have included genetic perturbation tasks in downstream evaluations to assess their applicability to perturbation modeling~\cite{cui2024scgpt, hao2024large}. Among these, scGPT performs large-scale generative pretraining on single-cell transcriptomic data and subsequently applies the learned representations to perturbation response prediction, treating perturbations as conditional inputs during downstream modeling~\cite{cui2024scgpt}. Similarly, scFoundation is pretrained on large collections of single-cell data and evaluated across a range of downstream tasks, including genetic perturbation settings, demonstrating the applicability of pretrained representations in perturbation-related analyses~\cite{hao2024large}. Beyond within-species modeling, GeneCompass incorporates cross-species information and external biological knowledge into a unified framework and evaluates its representations on perturbation-related tasks, highlighting the potential of cross-species pretraining~\cite{yang2024genecompass}. In addition, several approaches leverage pretrained language model embeddings for perturbation prediction. Methods such as GenePert, scouter, and scELMo leverage pretrained representations as gene- or cell-level features for perturbation prediction~\cite{chen2024genepert, zhu2025scouter, liu2025scelmo}, while recent work, including Geneformer and LPM, further explores large-scale pretraining for learning general-purpose representations applicable to perturbation-related tasks~\cite{cui2024geneformer, miladinovic2025silico}. These models are typically pretrained on unperturbed single-cell expression data, with genetic perturbation incorporated at the downstream stage. CytoWave further introduces a perturbation-centric pretraining framework that learns perturbation-aware representations by modeling perturbation-dependent changes in latent cellular states.

\subsection{Genetic Perturbation Models}
Many existing methods for single-cell perturbation modeling are formulated as dataset-specific supervised models, typically trained and evaluated within a single dataset and experimental setting~\cite{lotfollahi2023predicting, roohani2024predicting, bai2024attentionpert, chen2024genepert, ahlmann2024deep}. CPA introduces a compositional representation in latent space, where perturbation responses are factorized and combined to support prediction under complex and combinatorial perturbation settings~\cite{lotfollahi2023predicting}. GEARS focuses on generalizing to unseen multi-gene perturbations by incorporating gene interaction networks to capture dependencies among perturbed genes~\cite{roohani2024predicting}. AttentionPert models interactions among perturbed genes through attention mechanisms, enabling it to capture joint effects under multiplexed perturbations~\cite{bai2024attentionpert}. In addition to these deep learning approaches, simple baselines including linear models and mean-based predictors are commonly included in perturbation benchmarks, and recent studies have evaluated their performance alongside more complex models under diverse settings~\cite{ahlmann2024deep, vinas2025systema}. Beyond direct prediction of expression responses, another line of work models perturbation responses at the distribution level. Optimal transport-based methods learn population-level mappings between control and perturbed cells~\cite{bunne2023learning}. Flow-based approaches model population-level response distributions in expression space~\cite{klein2025cellflow}. Probabilistic methods rely on latent variable formulations to describe perturbation responses and provide uncertainty-aware representations~\cite{palla2025scalable}. These approaches offer complementary perspectives by modeling perturbation-dependent changes beyond pointwise prediction. 
Building on these advances, CytoWave extends the modeling framework by introducing perturbation-centric pretraining across diverse datasets and condition combinations. Instead of focusing on a single experimental setting, it incorporates diverse large-scale perturbation data to learn perturbation-aware representations, improving robustness and generalization across various biological contexts.

\section{Methods}
\subsection{Problem Formulation}

Let $\mathcal{G}$ denote a gene vocabulary with $|\mathcal{G}|=G$ genes,
and let $\mathcal{C}$ denote a shared vocabulary of categorical metadata tokens
(e.g., cell line, species, perturbation technology, and other experimental conditions).

We study large-scale pretraining for modeling single-cell genetic perturbation responses
from a unified perturbation corpus.
Each perturbation example is represented as a tuple
$(\mathbf{x}_i, \mathbf{y}_i, \mathbf{u}_i, \mathbf{c}_i)$,
where $\mathbf{x}_i \in \mathbb{R}_{\ge0}^{G}$ is a control expression profile,
$\mathbf{y}_i \in \mathbb{R}_{\ge0}^{G}$ is a perturbed expression profile
measured under the same experimental condition.
Control cells within the same experiment are treated as a reference population for perturbed cells under matched conditions. 
For each perturbed cell, a reference control profile is sampled from control cells sharing the same experimental context and metadata group (e.g., cell type and stimulation condition). These reference controls are used to define baseline cellular states for perturbation response modeling and evaluation.
$\mathbf{u}_i \subseteq \mathcal{G}$ is the set of perturbed genes (single- or multi-gene),
and $\mathbf{c}_i = (c_{i,1}, \ldots, c_{i,S_i})$ is a sequence of metadata tokens.

The objective of pretraining is to learn transferable representations of cellular state
and perturbation responses that generalize across datasets, biological conditions,
and perturbation combinations.
For downstream perturbation response prediction,
the model takes $(\mathbf{x}_i, \mathbf{u}_i, \mathbf{c}_i)$ as input
and predicts the post-perturbation expression $\mathbf{y}_i$. For evaluation, perturbation responses are measured as expression changes relative to the control profile.
This formulation aims to learn perturbation responses that generalize across diverse cellular contexts, datasets, and perturbation combinations.
% ============================================================
\subsection{CytoWave Architecture}
The model is implemented as a Transformer-based encoder–operator–decoder architecture. This design decomposes perturbation modeling into encoding cellular states, modeling perturbation-dependent changes, and decoding gene-level responses.
As shown in \Cref{fig:overall_pipeline}, CytoWave consists of a shared encoder $E(\cdot)$ that maps expression profiles and metadata to a latent cell state $\mathbf{z}\in\mathbb{R}^d$, a perturbation embedder $S(\cdot)$ that encodes perturbation sets $\mathbf{u}_i$, a perturbation operator $T(\cdot)$ that models perturbation-dependent changes in latent cellular states, and a query-based decoder $D(\cdot)$ that predicts gene expression values from latent states.

\paragraph{Input tokenization.}
A unified vocabulary is defined over genes and metadata tokens, denoted by $\mathcal{G}$ and $\mathcal{C}$, respectively.
To enable consistent modeling across species and experimental conditions, genes are mapped to orthology groups, which group homologous genes with conserved functions across species, providing a stable and biologically grounded basis for representing perturbations across datasets.

All tokens are embedded using a shared lookup table
$\mathrm{Emb}:\mathcal{G}\cup\mathcal{C}\rightarrow\mathbb{R}^d$,
together with a learned role embedding
$\mathrm{Emb}_{r}:\mathcal{R}\rightarrow\mathbb{R}^d$
to distinguish their semantic roles (gene, perturbation, or metadata).
For a token $t$, the base embedding is given by
\begin{equation}
\mathbf{h}^{(0)}_t = \mathrm{Emb}(t) + \mathrm{Emb}_{r}(r_t).
\end{equation}

Gene expression values are log-normalized as
$\mathbf{v}\leftarrow \log(1+\mathbf{v}^{\mathrm{raw}})$.
To improve computational efficiency and reduce sensitivity to sequencing depth variation and low-abundance noise,
a sparse input is adopted by retaining only expressed genes,
\begin{equation}
\Omega(\mathbf{v})=\{g\in\mathcal{G}\mid v_g>0\}.
\end{equation}

To preserve continuous expression magnitudes that are important for modeling fine-grained perturbation responses, expression values are incorporated via FiLM~\citep{perez2018film}.
For an expressed gene $g\in\Omega(\mathbf{v}_i)$ with value $v_{i,g}$, a value-conditioned token is constructed as
\begin{gather}
(\boldsymbol{\gamma}_{i,g},\boldsymbol{\beta}_{i,g}) = f_{\mathrm{FiLM}}(v_{i,g}), \\
\mathbf{h}^{(0)}_{i,g}
=
\boldsymbol{\gamma}_{i,g}\odot \mathrm{Emb}(g)
+
\boldsymbol{\beta}_{i,g}
+
\mathrm{Emb}_{r}(r^{\mathrm{expr}}),
\end{gather}
where $f_{\mathrm{FiLM}}$ is a small MLP and $\odot$ denotes element-wise multiplication.

The resulting tokens are input into the Transformer encoder without positional encodings, ensuring permutation invariance consistent with the unordered nature of transcriptomic data.

\paragraph{Transformer encoder and cell state aggregation.}
For each cell $i$, FiLM-conditioned gene tokens and metadata tokens are combined into a single set
\begin{equation}
\mathcal{H}^{(0)}_i = \{\mathbf{h}^{(0)}_{i,s}\}_{s=1}^{S_i} \cup \{\mathbf{h}^{(0)}_{i,g}\}_{g\in\Omega(\mathbf{v}_i)}.
\end{equation}
A stack of $L$ Transformer encoder layers processes this unordered set without positional encodings, producing contextualized token states $\mathcal{H}^{(L)}_i$ that capture gene--gene and gene--metadata interactions.

The latent cell state is obtained by aggregating the contextualized token states via mean pooling:
\begin{equation}
\mathbf{z}_i = E(\mathbf{v}_i,\mathbf{c}_i)
= \frac{1}{|\mathcal{H}^{(L)}_i|}\sum_{\mathbf{h}\in\mathcal{H}^{(L)}_i}\mathbf{h} \in \mathbb{R}^d.
\end{equation}

\paragraph{Perturb signal representation and operator.}
A (multi-)gene perturbation set is embedded using a permutation-invariant mapping $S(\cdot)$, implemented as:
\begin{equation}
\mathbf{s}_i=S(\mathbf{u}_i)=\frac{1}{|\mathbf{u}_i|}\sum_{g\in\mathbf{u}_i}\left(\mathrm{Emb}(g)+\mathrm{Emb}_{r}(r^{\mathrm{expr}})\right).
\end{equation}

The perturbation operator models perturbation-dependent changes in latent cellular states.
Given the control cell state $\mathbf{z}^{(x)}_i = E(\mathbf{x}_i,\mathbf{c}_i)$, it predicts perturbation-dependent latent state changes and applies it additively:
\begin{equation}
\tilde{\mathbf{z}}^{(y)}_i
= T(\mathbf{z}^{(x)}_i,\mathbf{s}_i)
= \mathbf{z}^{(x)}_i + t_\theta([\mathbf{z}^{(x)}_i;\mathbf{s}_i]),
\end{equation}
where $t_\theta$ is an MLP and $[\cdot;\cdot]$ denotes concatenation.

\paragraph{Query-based decoder.}
To predict gene expression from latent cell states, we adopt a query-based decoding scheme.
Given a latent state $\mathbf{z}$ and a target gene $g$, a gene-specific query is formed as
\begin{equation}
\mathbf{q}=\mathbf{z}+\mathrm{Emb}(g)+\mathrm{Emb}_{r}(r^{\mathrm{expr}}),
\end{equation}
which is mapped to a scalar prediction:
\begin{equation}
\hat{v}_{g} = D(\mathbf{q}).
\end{equation}

This formulation enables flexible prediction over arbitrary subsets of genes and naturally supports both masked recovery and perturbation-conditioned expression prediction.

\paragraph{State Change--Perturbation Alignment Head.} We define $\mathbf{z}^{(\mathrm{meta})}(\mathbf{c}_i)$ as the mean-pooled representation of metadata tokens for cell $i$.
To align perturbation signals with latent state changes, we encode both control and perturbed profiles and define the latent state change:
\begin{equation}
\mathbf{z}^{(x)}_i = E(\mathbf{x}_i,\mathbf{c}_i),\;
\mathbf{z}^{(y)}_i = E(\mathbf{y}_i,\mathbf{c}_i),\;
\Delta\mathbf{z}_i = \mathbf{z}^{(y)}_i - \mathbf{z}^{(x)}_i.
\end{equation}

During training, perturbation annotations are partially masked to obtain an observed subset $\mathbf{u}^{\mathrm{obs}}_i$.
The alignment head maps the latent state change together with contextual information to a predicted perturbation embedding:
\begin{equation}
\hat{\mathbf{e}}_i = R_\psi\!\left([\Delta\mathbf{z}_i;\ \mathbf{z}^{(\mathrm{meta})}(\mathbf{c}_i) + S(\mathbf{u}^{\mathrm{obs}}_i)]\right)\in\mathbb{R}^d.
\end{equation}

% ============================================================
\subsection{Training Objectives}
% ============================================================
CytoWave is pretrained using a two-stage framework.
In Stage~1, the model learns representations of cellular state by reconstructing masked gene expression from control cells.
In Stage~2, the model learns perturbation-dependent state changes between reference control states and perturbed cellular states, with bidirectional objectives that combine direct perturbation response prediction and reverse perturbation signal alignment. Our goal is to learn transferable representations of cellular state and perturbation signals from a unified perturbation corpus, enabling generalization across datasets, conditions, and perturbation combinations.

% ----------------------------
\subsubsection{Stage 1: Reference Cell State Modeling via Masked Recovery}
Stage~1 learns perturbation-agnostic representations of cellular state from control profiles, providing a reference baseline for modeling perturbation-dependent changes. This is achieved via masked gene-value recovery on control cells.

Given a control expression profile $\mathbf{x}_i$, a subset of expressed genes $\mathcal{M}_i^{(x)}\subseteq \Omega(\mathbf{x}_i)$ is randomly sampled with ratio $\rho_x$ and masked to form a corrupted input $\tilde{\mathbf{x}}_i$.
For each masked gene $g\in\mathcal{M}_i^{(x)}$, we predict
\begin{equation}
\hat{x}_{i,g}
=
D\!\left(E(\tilde{\mathbf{x}}_i,\mathbf{c}_i)+\mathrm{Emb}(g)+\mathrm{Emb}_{r}(r^{\mathrm{expr}})\right).
\end{equation}
The model is trained to minimize mean-squared error over masked entries:
\begin{equation}
\mathcal{L}_{\mathrm{ctrl}}=
\mathbb{E}_i\left[\frac{1}{|\mathcal{M}_i^{(x)}|}\sum_{g\in\mathcal{M}_i^{(x)}}\left(\hat{x}_{i,g}-x_{i,g}\right)^2\right].
\end{equation}
% ----------------------------
\subsubsection{Stage 2: Bidirectional Learning of Perturbation-Dependent Latent State Changes}
\label{sec:stage2}

Stage~2 models perturbation-dependent changes between reference control states and perturbed cellular states.
The model predicts post-perturbation gene expression conditioned on control states and perturbation signals.
Direct prediction captures the resulting expression patterns.
To further couple learned latent state changes with perturbation signals, we introduce a bidirectional learning scheme
that combines direct perturbation response prediction (Stage~2A) with inverse perturbation signal alignment via masked perturbation tokens (Stage~2B).
This design encourages the learned latent state changes to remain grounded in perturbation signals.
% ----------------------------

\paragraph{Stage 2A: Direct Perturbation Response Prediction}
Training pairs are constructed by sampling control cells from the same experimental condition and matched metadata group for each perturbed cell. Given $(\mathbf{x}_i,\mathbf{y}_i,\mathbf{u}_i,\mathbf{c}_i)$, we compute
$\mathbf{z}^{(x)}_i = E(\mathbf{x}_i,\mathbf{c}_i)$,
$\mathbf{s}_i=S(\mathbf{u}_i)$, and
$\tilde{\mathbf{z}}^{(y)}_i=T(\mathbf{z}^{(x)}_i,\mathbf{s}_i)$.
Supervision is applied on a sampled query set $\mathcal{Q}_i\subseteq\mathcal{G}$ with ratio $\rho_q$, where
\begin{equation}
\hat{y}_{i,g}=D\!\left(\tilde{\mathbf{z}}^{(y)}_i+\mathrm{Emb}(g)+\mathrm{Emb}_{r}(r^{\mathrm{expr}})\right),\quad g\in\mathcal{Q}_i.
\end{equation}
The loss is
\begin{equation}
\mathcal{L}_{\mathrm{expr}} =
\mathbb{E}_i\left[\frac{1}{|\mathcal{Q}_i|}\sum_{g\in\mathcal{Q}_i}(\hat{y}_{i,g}-y_{i,g})^2\right].
\end{equation}

\paragraph{Stage 2B: Reverse Perturbation Signal Alignment}
The perturbation set is split into $\mathbf{u}^{\mathrm{obs}}_i$ and $\mathbf{u}^{\mathrm{mask}}_i$ by masking each perturbation with probability $\rho_u$, ensuring $\mathbf{u}^{\mathrm{mask}}_i\neq\emptyset$, and defining $\mathbf{e}^{+}_i = S(\mathbf{u}^{\mathrm{mask}}_i)$.
We compute the latent state change
\begin{equation}
\mathbf{z}^{(y)}_i = E(\mathbf{y}_i,\mathbf{c}_i),\quad
\Delta \mathbf{z}_i = \mathbf{z}^{(y)}_i - \mathbf{z}^{(x)}_i,
\end{equation}
and predict perturbation embeddings
\begin{equation}
\hat{\mathbf{e}}_i =
R_\psi\!\left(
[\Delta \mathbf{z}_i;\ \mathbf{z}^{(\mathrm{meta})}(\mathbf{c}_i) + S(\mathbf{u}^{\mathrm{obs}}_i)]
\right).
\end{equation}
Alignment is enforced using an in-batch InfoNCE loss, where $\mathrm{sim}(\cdot,\cdot)$ denotes cosine similarity and $\tau=0.1$ is the temperature hyperparameter:
\begin{equation}
\mathcal{L}_{\mathrm{align}} =
\mathbb{E}_{i\in\mathcal{B}}
\left[
-\log
\frac{\exp\left(\mathrm{sim}(\hat{\mathbf{e}}_i,\mathbf{e}^{+}_i)/\tau\right)}
{\sum_{j\in\mathcal{B}}\exp\left(\mathrm{sim}(\hat{\mathbf{e}}_i,\mathbf{e}^{+}_j)/\tau\right)}
\right].
\end{equation}

\paragraph{Stage 2 Overall Objective}
The direct perturbation response prediction and reverse perturbation signal alignment objectives are jointly optimized:
\begin{equation}
\mathcal{L}_{\mathrm{stage2}} = \mathcal{L}_{\mathrm{expr}} + \lambda_{\mathrm{align}}\,\mathcal{L}_{\mathrm{align}}.
\end{equation}
% ============================================================

\section{Experiments}

\subsection{Datasets}
To support large-scale, cross-condition, and cross-species modeling of single-cell genetic perturbation responses, we provide a unified orthology-aligned pretraining corpus curated from 19 independent studies~\cite{Zhu2025.12.23.696273, datlinger2017pooled, datlinger2021ultra, dixit2016perturb, frangieh2021multimodal, gasperini2019genome, lara2023vivo, liang2023organoid, nadig2025transcriptome, norman2019exploring, papalexi2021characterizing, replogle2022mapping, santinha2023transcriptional, shifrut2018genome, tian2019crispr, tian2021genome, wessels2023efficient, xie2017multiplexed, xu2024dissecting}. The resulting corpus comprises 42 datasets from human and mouse experiments, totaling approximately 24.7 million single-cell transcriptomes.

\setlength{\columnsep}{15pt}
\setlength{\intextsep}{5pt}
\begin{wrapfigure}{l}{0.4\textwidth}
  \centering
  \includegraphics[width=0.4\textwidth]{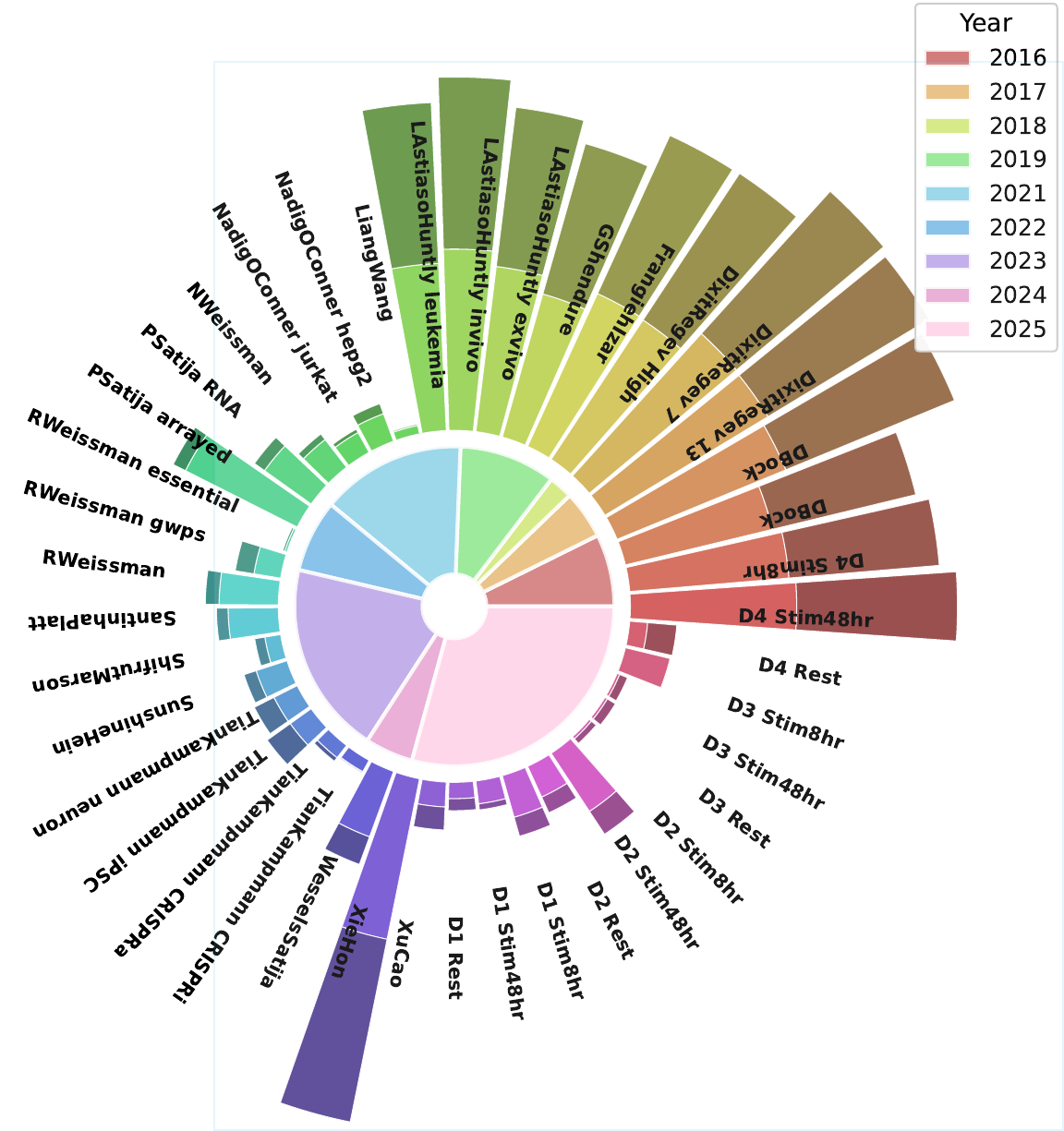}
  \caption{Overview of datasets collected for CytoWave.
Bar lengths represent the number of control cells (lighter color) and perturbed cells (darker color) in each dataset.
The central marker indicates the year of dataset publication (2016--2025).}
  \label{fig:dataset}
\end{wrapfigure}

We show the dataset year distribution and the proportion of control versus perturbed cells in Figure~\ref{fig:dataset}. All datasets are mapped to orthology identifiers and represented in a unified sparse format with consistent gene and perturbation vocabularies. Each cell is annotated with a standardized set of metadata tokens describing species, cellular context, cell line, lineage, perturbation technology, disease state, and stimulation conditions, with control cells serving as baseline references within each experimental condition. Details on data collection and statistics are given in the Appendices~\ref{app:data} and~\ref{app:dataset_stats}.

From the curated corpus, we construct a downstream benchmark to evaluate cross-dataset generalization. 
We adopt a dataset-held-out protocol: five representative datasets are excluded from pretraining and used exclusively for evaluation. 
These datasets cover diverse experimental conditions, including different species, CRISPR technologies, cell types, and biological contexts~\cite{datlinger2017pooled, gasperini2019genome, papalexi2021characterizing, santinha2023transcriptional}. 
CytoWave is pretrained on the remaining datasets, allowing us to assess whether learned perturbation representations transfer to unseen condition combinations.
After pretraining, CytoWave is finetuned on the training split of each held-out downstream dataset and evaluated on the corresponding test split.

\subsection{Baselines}
CytoWave is compared with a diverse set of baselines covering different modeling paradigms. 
Simple statistical baselines include Linear~\cite{ahlmann2024deep} and TrainMean, while basic machine learning baselines include BaseReg and BaseMLP~\cite{wei2025benchmarking}. 
Supervised perturbation-specific models include CPA~\cite{lotfollahi2023predicting} and GEARS~\cite{roohani2024predicting}. Pretrained single-cell models are also evaluated on perturbation tasks. scGPT~\cite{cui2024scgpt} is included to represent the finetuning performance of single-cell pretrained models. All pretrained models, including CytoWave, are finetuned on the downstream training split before evaluation. Representation-based transfer methods include GenePert~\cite{chen2024genepert} and scouter~\cite{zhu2025scouter}. Recently proposed generative and pretrained frameworks, including scLDM~\cite{palla2025scalable}, LPM~\cite{miladinovic2025silico}, and GeneFormer~\cite{cui2024geneformer}, are additionally included to provide a comprehensive comparison across modeling paradigms.

\subsection{Results}
CytoWave is evaluated on gene-level perturbation response prediction. Performance is assessed using three complementary metrics. $\Delta$PCC measures the correlation between predicted and ground-truth expression changes. $\Delta$PCC$_{20}$ focuses on the most responsive genes. MAE quantifies the magnitude of prediction errors over genes with measurable expression activity in either the control or perturbed profile. A pathway-level analysis based on AUCell is included as a biological case study to examine whether predicted responses recover known pathway activities. An ablation study analyzes the contributions of different pretraining stages and metadata inputs. Results are averaged over three random seeds (0, 1, and 2). All methods are evaluated under the same cell-level protocol for consistency. Deterministic population-level predictors may therefore be conservative under single-cell $\Delta$PCC evaluation.

\begin{table*}[t]
\centering
\tiny
\setlength{\tabcolsep}{3.5pt}
\caption{Single-gene perturbation prediction across all datasets.
We report MAE (lower is better) together with $\Delta$PCC and $\Delta$PCC$_{20}$ (higher is better).}
\label{tab:all_metrics}

\resizebox{\textwidth}{!}{
\begin{tabular}{lccccc}
\toprule
\multicolumn{6}{c}{\textbf{MAE (lower is better)}} \\
\midrule
Method & DatlingerBock2017 & DatlingerBock2017(Stim) & PapalexiSatija2021 & GasperiniShendure2019 & SantinhaPlatt2023 \\
\midrule
BaseReg & 2.016 & 1.996 & 2.200 & 2.547 & 2.057 \\
Linear & 2.038 & 2.003 & 2.200 & 2.627 & 2.057 \\
TrainMean & 2.007 & 2.000 & 2.214 & 2.519 & 2.057 \\
BaseMLP & 2.017 $\pm$ 0.015 & 1.976 $\pm$ 0.030 & 2.314 $\pm$ 0.006 & 2.632 $\pm$ 0.031 & 2.134 $\pm$ 0.014 \\
CPA & \underline{1.080 $\pm$ 0.009} & 0.982 $\pm$ 0.008 & 0.997 $\pm$ 0.004 & 1.332 $\pm$ 0.005 & 1.001 $\pm$ 0.004 \\
GEARS & 1.845 $\pm$ 0.033 & 1.224 $\pm$ 0.025 & 1.354 $\pm$ 0.024 & 1.738 $\pm$ 0.018 & 1.258 $\pm$ 0.009 \\
scGPT & 1.883 $\pm$ 0.021 & 1.698 $\pm$ 0.037 & 1.407 $\pm$ 0.044 & 2.145 $\pm$ 0.011 & 1.986 $\pm$ 0.075 \\
GenePert & 2.001 $\pm$ 0.002 & 2.016 $\pm$ 0.003 & 2.223 $\pm$ 0.003 & 3.454 $\pm$ 0.079 & 2.059 $\pm$ 0.001 \\
scouter & 2.032 $\pm$ 0.012 & 2.012 $\pm$ 0.010 & 2.628 $\pm$ 0.011 & 2.541 $\pm$ 0.009 & 2.151 $\pm$ 0.007 \\
scLDM & 1.229 $\pm$ 0.054 & 1.162 $\pm$ 0.047 & 1.014 $\pm$ 0.026 & 1.355 $\pm$ 0.019 & 1.575 $\pm$ 0.030 \\
LPM & 1.184 $\pm$ 0.016 & 1.421 $\pm$ 0.035 & 1.115 $\pm$ 0.018 & 1.787 $\pm$ 0.025 & 1.231 $\pm$ 0.025 \\
GeneFormer & 1.210 $\pm$ 0.043 & 1.333 $\pm$ 0.024 & 1.584 $\pm$ 0.014 & 1.688 $\pm$ 0.045 & 1.258 $\pm$ 0.014 \\
\midrule
CytoWave-S1 & 1.082 $\pm$ 0.011 & \underline{0.891 $\pm$ 0.012} & \underline{0.852 $\pm$ 0.018} & \underline{1.298 $\pm$ 0.005} & \underline{0.871 $\pm$ 0.018} \\
CytoWave-w/o Meta & 1.144 $\pm$ 0.020 & 0.921 $\pm$ 0.007 & 0.885 $\pm$ 0.021 & 1.317 $\pm$ 0.007 & 0.957 $\pm$ 0.022 \\
CytoWave-S2 & \textbf{1.077 $\pm$ 0.031} & \textbf{0.867 $\pm$ 0.009} & \textbf{0.791 $\pm$ 0.006} & \textbf{1.133 $\pm$ 0.009} & \textbf{0.833 $\pm$ 0.011} \\
\midrule

\multicolumn{6}{c}{\textbf{$\Delta$PCC (higher is better)}} \\
\midrule
BaseReg & -0.075 & 0.002 & 0.181 & 0.037 & 0.006 \\
Linear & -0.059 & 0.019 & 0.119 & 0.024 & 0.014 \\
TrainMean & -0.097 & -0.148 & 0.223 & 0.026 & 0.008 \\
BaseMLP & 0.026 $\pm$ 0.027 & 0.098 $\pm$ 0.032 & 0.213 $\pm$ 0.008 & 0.057 $\pm$ 0.002 & 0.058 $\pm$ 0.002 \\
CPA & 0.132 $\pm$ 0.020 & 0.237 $\pm$ 0.025 & 0.251 $\pm$ 0.021 & 0.102 $\pm$ 0.017 & 0.101 $\pm$ 0.011 \\
GEARS & 0.012 $\pm$ 0.013 & 0.002 $\pm$ 0.005 & -0.010 $\pm$ 0.006 & -0.001 $\pm$ 0.008 & 0.006 $\pm$ 0.004 \\
scGPT & 0.026 $\pm$ 0.004 & 0.034 $\pm$ 0.007 & 0.042 $\pm$ 0.010 & 0.036 $\pm$ 0.006 & 0.003 $\pm$ 0.001 \\
GenePert & -0.009 $\pm$ 0.010 & -0.028 $\pm$ 0.010 & 0.056 $\pm$ 0.013 & -0.018 $\pm$ 0.015 & -0.013 $\pm$ 0.004 \\
scouter & -0.031 $\pm$ 0.012 & 0.018 $\pm$ 0.016 & 0.182 $\pm$ 0.013 & 0.066 $\pm$ 0.001 & 0.025 $\pm$ 0.004 \\
scLDM & 0.008 $\pm$ 0.001 & 0.006 $\pm$ 0.001 & 0.005 $\pm$ 0.002 & 0.031 $\pm$ 0.004 & 0.055 $\pm$ 0.008 \\
LPM & -0.035 $\pm$ 0.002 & 0.153 $\pm$ 0.001 & 0.173 $\pm$ 0.029 & 0.035 $\pm$ 0.003 & 0.100 $\pm$ 0.013 \\
GeneFormer & 0.001 $\pm$ 0.000 & 0.011 $\pm$ 0.002 & 0.012 $\pm$ 0.001 & 0.001 $\pm$ 0.001 & 0.005 $\pm$ 0.001 \\
\midrule
CytoWave-S1 & \underline{0.187 $\pm$ 0.021} & \underline{0.278 $\pm$ 0.012} & \underline{0.252 $\pm$ 0.008} & \underline{0.108 $\pm$ 0.001} & \textbf{0.108 $\pm$ 0.008} \\
CytoWave-w/o Meta & 0.185 $\pm$ 0.018 & 0.268 $\pm$ 0.014 & 0.250 $\pm$ 0.019 & 0.107 $\pm$ 0.005 & 0.102 $\pm$ 0.009 \\
CytoWave-S2 & \textbf{0.189 $\pm$ 0.008} & \textbf{0.279 $\pm$ 0.010} & \textbf{0.255 $\pm$ 0.014} & \textbf{0.112 $\pm$ 0.005} & \underline{0.107 $\pm$ 0.004} \\
\midrule

\multicolumn{6}{c}{\textbf{$\Delta$PCC$_{20}$ (higher is better)}} \\
\midrule
BaseReg & -0.079 & -0.076 & 0.200 & 0.078 & 0.028 \\
Linear & -0.044 & -0.026 & 0.133 & 0.056 & 0.019 \\
TrainMean & -0.136 & -0.169 & 0.135 & 0.055 & 0.038 \\
BaseMLP & 0.016 $\pm$ 0.025 & 0.060 $\pm$ 0.015 & 0.243 $\pm$ 0.001 & 0.084 $\pm$ 0.003 & 0.130 $\pm$ 0.026 \\
CPA & 0.261 $\pm$ 0.003 & \underline{0.328 $\pm$ 0.001} & \underline{0.304 $\pm$ 0.004} & 0.149 $\pm$ 0.009 & 0.246 $\pm$ 0.003 \\
GEARS & -0.014 $\pm$ 0.012 & 0.018 $\pm$ 0.017 & -0.014 $\pm$ 0.023 & -0.004 $\pm$ 0.020 & 0.002 $\pm$ 0.015 \\
scGPT & 0.047 $\pm$ 0.010 & 0.053 $\pm$ 0.009 & 0.084 $\pm$ 0.011 & 0.047 $\pm$ 0.005 & 0.072 $\pm$ 0.013 \\
GenePert & -0.022 $\pm$ 0.010 & -0.057 $\pm$ 0.003 & 0.079 $\pm$ 0.020 & -0.039 $\pm$ 0.022 & 0.014 $\pm$ 0.006 \\
scouter & 0.160 $\pm$ 0.009 & 0.214 $\pm$ 0.006 & 0.232 $\pm$ 0.013 & 0.139 $\pm$ 0.004 & 0.113 $\pm$ 0.006 \\
scLDM & -0.065 $\pm$ 0.011 & 0.010 $\pm$ 0.022 & 0.048 $\pm$ 0.008 & 0.139 $\pm$ 0.050 & 0.114 $\pm$ 0.020 \\
LPM & -0.003 $\pm$ 0.011 & 0.095 $\pm$ 0.014 & 0.296 $\pm$ 0.018 & 0.056 $\pm$ 0.004 & 0.200 $\pm$ 0.018 \\
GeneFormer & -0.009 $\pm$ 0.007 & -0.020 $\pm$ 0.003 & 0.220 $\pm$ 0.003 & 0.006 $\pm$ 0.008 & 0.175 $\pm$ 0.005 \\
\midrule
CytoWave-S1 & \underline{0.267 $\pm$ 0.020} & 0.300 $\pm$ 0.008 & 0.302 $\pm$ 0.008 & 0.142 $\pm$ 0.003 & 0.243 $\pm$ 0.011 \\
CytoWave-w/o Meta & 0.265 $\pm$ 0.025 & 0.313 $\pm$ 0.011 & 0.282 $\pm$ 0.023 & \underline{0.150 $\pm$ 0.007} & \underline{0.248 $\pm$ 0.019} \\
CytoWave-S2 & \textbf{0.272 $\pm$ 0.006} & \textbf{0.330 $\pm$ 0.002} & \textbf{0.305 $\pm$ 0.001} & \textbf{0.151 $\pm$ 0.005} & \textbf{0.251 $\pm$ 0.005} \\
\bottomrule
\end{tabular}
}
\end{table*}
\subsubsection{Perturbation Response Prediction}
Table~\ref{tab:all_metrics} summarizes genetic perturbation prediction performance across five evaluation datasets, reported using $\Delta$PCC$_{20}$, $\Delta$PCC, and MAE. 
CytoWave-S2 consistently achieves the strongest performance across all metrics, demonstrating stable improvements across datasets. 
It shows robust performance across diverse regimes, including both unstimulated and stimulated conditions (DatlingerBock2017), datasets with dense cellular coverage (PapalexiSatija2021), and large-scale perturbation screens with many targets (GasperiniShendure2019). 
Strong results are also observed on SantinhaPlatt2023, indicating effective generalization across species and experimental settings. Overall, these results suggest that CytoWave provides a reliable and consistent solution for predicting genetic perturbation responses across different species and experimental conditions.

\subsubsection{Case Study: Pathway Activity}

\begin{figure*}[t!]
  \centering
  \includegraphics[width=\textwidth]{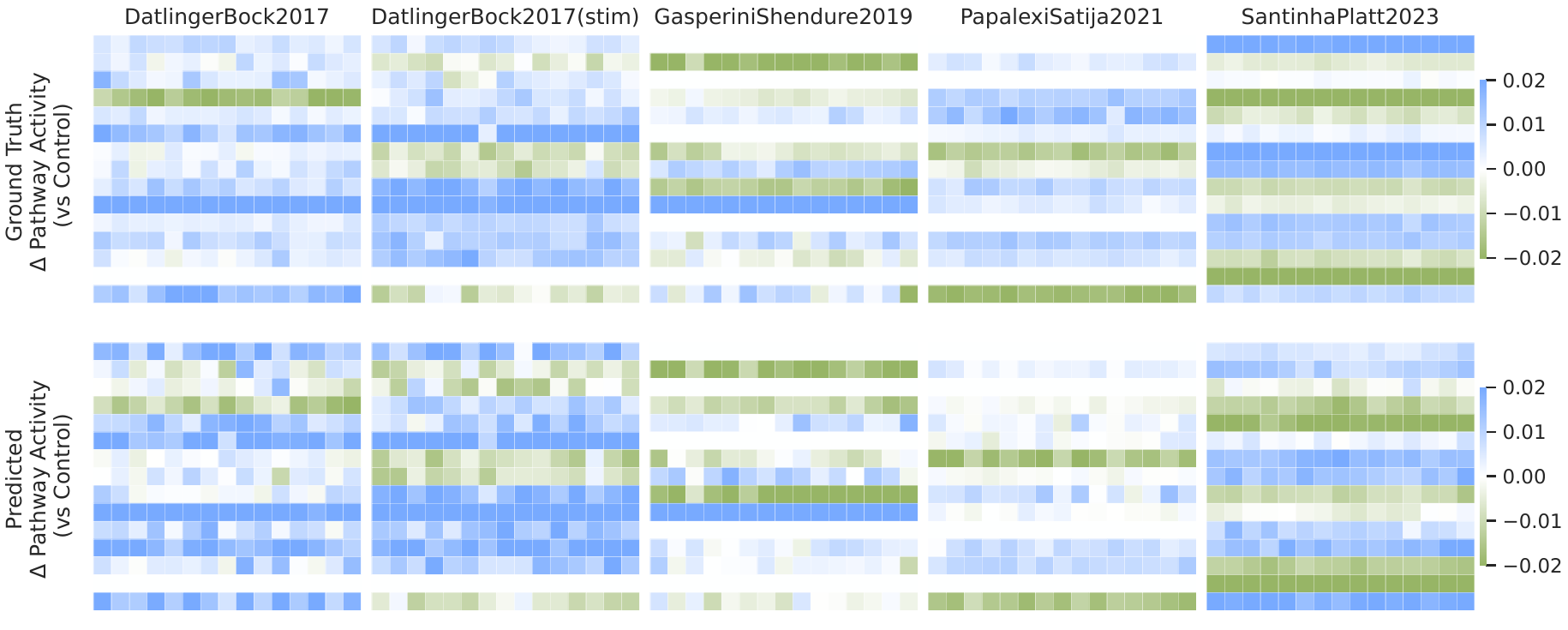}
  \caption{
    Ground-truth and predicted pathway activity changes across multiple single-cell perturbation datasets. An illustrative subset of high-agreement pathways and perturbations is shown. Top panels denote ground truth, and bottom panels denote predictions.
    }
  \label{fig:pathway}
\end{figure*}

To assess whether CytoWave captures biologically meaningful perturbation responses beyond gene-level expression, we examine pathway activity using AUCell scores based on the Hallmark 50 gene sets. 
AUCell scores are computed on both ground-truth and predicted expression profiles to obtain pathway-level representations.
Figure~\ref{fig:pathway} shows pathway activity patterns across five benchmark datasets. 
CytoWave recovers consistent activation and repression patterns associated with perturbations, with strong agreement between predicted and ground-truth heatmaps. 
This agreement is observed across different experimental settings, suggesting that the model captures coherent biological responses at the pathway level.
Overall, the alignment of pathway activity patterns provides interpretable evidence that CytoWave captures structured perturbation responses beyond individual gene-level predictions. 
Additional details and extended results are provided in Appendix~\ref{app:pathway}.

\subsubsection{Ablation Study}
We conduct ablation studies to analyze the contribution of different components in CytoWave, including Stage~1 pretraining, Stage~2 perturbation-aware learning, and metadata inputs. CytoWave-S1 already achieves strong performance, indicating that large-scale pretraining on control cells provides a solid foundation for modeling transcriptional states. 
Building on this, CytoWave-S2 consistently improves performance across all evaluation metrics, demonstrating the benefit of incorporating perturbation-aware learning. 
Additional results further show that metadata contributes to performance while not being the sole driver of improvements. Overall, these results highlight the complementary roles of large-scale pretraining and perturbation-aware learning in improving prediction performance.

\section{Conclusion}
We present CytoWave, a large-scale perturbation-centric pretraining framework for single-cell genetic perturbation modeling. By unifying cross-species perturbation datasets through orthology alignment and adopting a two-stage self-supervised pretraining strategy, CytoWave learns perturbation-aware representations that generalize across genes and experimental contexts. Across multiple benchmarks, CytoWave achieves strong and consistent performance, demonstrating the effectiveness of perturbation-centric pretraining. These results suggest that incorporating perturbation data during pretraining improves the modeling of transcriptional responses and supports the capture of structured biological response patterns. As large-scale perturbation datasets continue to grow in size, diversity, and experimental coverage, explicitly modeling perturbation responses during pretraining provides a promising direction for scalable and transferable representation learning in single-cell biology.

\section{Acknowledgement}
This work was partially supported by US National Science Foundation IIS-2412195, CCF-2400785, the Cancer Prevention and Research Institute of Texas (CPRIT) award (RP230363), the National Institutes of Health (NIH) R01 awards (1R01AI190103-01 and 1R01LM015497-01).

\bibliographystyle{unsrt}   
\bibliography{reference}  
% \section*{References}

% References follow the acknowledgments in the camera-ready paper. Use unnumbered first-level heading for
% the references. Any choice of citation style is acceptable as long as you are
% consistent. It is permissible to reduce the font size to \verb+small+ (9 point)
% when listing the references.
% Note that the Reference section does not count towards the page limit.
% \medskip

% {
% \small

% [1] Alexander, J.A.\ \& Mozer, M.C.\ (1995) Template-based algorithms for
% connectionist rule extraction. In G.\ Tesauro, D.S.\ Touretzky and T.K.\ Leen
% (eds.), {\it Advances in Neural Information Processing Systems 7},
% pp.\ 609--616. Cambridge, MA: MIT Press.

% [2] Bower, J.M.\ \& Beeman, D.\ (1995) {\it The Book of GENESIS: Exploring
%   Realistic Neural Models with the GEneral NEural SImulation System.}  New York:
% TELOS/Springer--Verlag.

% [3] Hasselmo, M.E., Schnell, E.\ \& Barkai, E.\ (1995) Dynamics of learning and
% recall at excitatory recurrent synapses and cholinergic modulation in rat
% hippocampal region CA3. {\it Journal of Neuroscience} {\bf 15}(7):5249-5262.
% }

%%%%%%%%%%%%%%%%%%%%%%%%%%%%%%%%%%%%%%%%%%%%%%%%%%%%%%%%%%%%
\newpage
\appendix
\begin{center}
    \Large \textbf{Appendix}
\end{center}
\section{Data and Code Availability}
We provide a detailed data curation and preprocessing protocol in Appendix~\ref{app:data}, and describe the implementation details in Appendix~\ref{app:impl}. All processed datasets and code will be made publicly available upon acceptance of the paper.

\section{Data Curation and Preprocessing}
\label{app:data}

This section describes the data curation, identifier harmonization, metadata normalization, and preprocessing pipeline used to construct the orthology-aligned corpus for CytoWave, with evaluation datasets held out from pretraining. All steps are applied uniformly across datasets to ensure reproducibility, cross-study consistency, and compatibility across species.

\subsection{Gene Identifier Normalization and Reference Mapping}

Gene identifiers in raw scRNA-seq datasets are provided using heterogeneous naming conventions, including inconsistent capitalization and isoform-specific suffixes. To standardize gene identities, gene symbols are first normalized by removing isoform suffixes and extraneous formatting.

Normalized gene symbols are mapped to species-specific Ensembl Gene IDs using organism-specific reference annotations. 
Human genes are mapped using the Human Gene Nomenclature Committee (HGNC) reference, while mouse genes are mapped using Mouse Genome Informatics (MGI) annotations together with Ensembl GRCm39 gene models. 
Only genes with valid Ensembl identifiers and orthology assignments are retained. This mapping establishes a stable and widely accepted identifier space for downstream processing.

To enable cross-species integration, Ensembl Gene IDs are further mapped to orthology-level identifiers using Ensembl Compara protein homology annotations (release~115). Gene models are defined using Ensembl reference genome annotations for human (GRCh38, Ensembl release~113) and mouse (GRCm39, Ensembl release~113). Each retained gene is thus associated with a unified ortholog identifier shared across species. Genes without a valid orthology assignment are excluded from the dataset.

\subsection{Perturbation Target Harmonization}

Perturbation annotations are harmonized using the same identifier and orthology references to ensure alignment between perturbation targets and gene expression features. Cells annotated as negative controls are assigned a dedicated \texttt{CONTROL} perturbation token.

For perturbed cells, target gene symbols are first normalized and then mapped to species-specific Ensembl Gene IDs using HGNC references for human datasets and MGI/Ensembl annotations for mouse datasets. Cross-species alignment is subsequently performed through Ensembl Compara protein homology annotations (release 115), which map species-specific Ensembl gene identifiers into a shared orthology space. Cells whose perturbation targets cannot be mapped to a valid ortholog identifier are removed. This procedure guarantees that all perturbations are represented in the same orthology-aligned token space as gene expression inputs.

\subsection{Metadata Standardization and Context Tokenization}

In addition to gene expression and perturbation labels, we curate a unified set of metadata tokens capturing biological and experimental context, including species, cell context, cell line, lineage, perturbation technology, disease status, stimulation condition, and stimulation time.

Metadata fields are normalized using rule-based string matching against predefined keyword dictionaries. Each metadata category is mapped to a controlled vocabulary of discrete tokens. When dataset-specific metadata fields are missing or ambiguous, predefined default tokens are assigned. This design yields a fixed-length metadata token representation per cell, independent of the original dataset schema.

\subsection{Cell and Gene Quality Control}

We apply several quality-control filters to remove low-quality or ambiguous observations:
\begin{itemize}
    \item Genes with zero expression across all cells are removed.
    \item Genes without valid orthology assignments are removed.
    \item Cells associated with unmapped perturbation targets are removed.
    \item Cells with fewer than 500 detected genes are excluded.
\end{itemize}

After filtering, all retained genes have valid orthology identifiers and non-zero expression in at least one cell, and all retained cells have valid perturbation annotations.

\subsection{Expression Normalization and Transformation}

Raw gene expression counts are normalized by library size on a per-cell basis. Specifically, counts are scaled such that the total expression per cell matches the median total count across the dataset. Median-based normalization reduces sensitivity to extreme library sizes while preserving relative expression magnitudes.

Normalized expression values are subsequently transformed using a $\log(1 + x)$ function to stabilize variance and compress dynamic range. Post-normalization expression distributions are inspected to confirm that total counts are centered around the dataset median and that log-transformed values fall within a stable numerical range.

\subsection{Pathway Gene Set Resources}

For pathway-level analyses, we use the Hallmark gene sets from the Molecular Signatures Database (MSigDB), release~2025.1. Human Hallmark gene sets are mapped to the same orthology space using Ensembl gene models (GRCh38, release~113) and Ensembl Compara homologies (release~115), enabling consistent pathway analysis across species.

\subsection{Preprocessing Validation and Provenance Tracking}

To ensure correctness and reproducibility, we perform systematic validation checks, including verification of reported versus computed numbers of detected genes per cell, verification of reported versus computed total counts per cell, and sanity checks on normalization scaling and log-transformed expression values.

All preprocessing steps, filtering statistics, and reference resource versions are recorded as structured metadata within each processed dataset. The resulting curated datasets are stored in standardized \texttt{.h5ad} formats and used as inputs for downstream sparse tokenization and model pretraining.

\subsection{Conversion to Foundation-Friendly Zarr Representation}
\label{app:zarr}

After preprocessing and curation in \texttt{.h5ad} format, we further convert each dataset into a foundation-model-oriented \texttt{zarr} representation. This conversion is motivated by the need to support scalable, distributed pretraining over tens of millions of cells while maintaining strict schema consistency and efficient access to sparse gene expression and tokenized metadata.

\paragraph{Motivation.}
While \texttt{.h5ad} provides a flexible container for exploratory analysis, it is not optimized for large-scale foundation model pretraining, which requires:
(i) streaming-friendly access patterns,
(ii) efficient storage of sparse expression matrices,
(iii) explicit separation between continuous expression values and discrete tokenized attributes, and
(iv) strict schema guarantees across heterogeneous datasets.
The \texttt{zarr} format satisfies these requirements by enabling chunked, compressed, and parallel I/O while remaining framework-agnostic.

\subsection{Expression Storage in CSR Form}

Gene expression matrices are stored in compressed sparse row (CSR) form to preserve sparsity and minimize storage overhead. Each dataset encodes expression values using three arrays corresponding to the CSR representation:
(i) nonzero expression values,
(ii) column indices (gene indices),
and (iii) row pointers defining cell boundaries.
The original matrix shape (number of cells and genes) is stored as metadata.

All expression values are stored in single-precision floating point format after normalization and log transformation. Dense matrices, when present, are converted to CSR prior to storage to enforce a uniform representation across datasets.

\subsection{Gene Token Representation}

Genes are represented exclusively by orthology-aligned tokens. For each dataset, a single ordered array defines the mapping from gene index to ortholog identifier. This array establishes the global gene vocabulary used during pretraining and ensures that gene identity is decoupled from dataset-specific ordering or naming conventions.

\subsection{Cell-Level Metadata Schema}

To enforce consistency across datasets, we define a fixed metadata schema consisting of the following token categories:
species, cell context, cell line, lineage, perturbation technology, disease status, stimulation condition, and stimulation time.
Each metadata field is stored as a per-cell array of discrete tokens, yielding a fixed-length metadata representation for every cell regardless of dataset origin.

Datasets missing specific metadata fields are assigned predefined default tokens during preprocessing, ensuring schema completeness. Any dataset failing to satisfy the required metadata schema is rejected during conversion.

\subsection{Perturbation Encoding}
\begin{figure}[t!]
  \centering
  \includegraphics[width=\textwidth]{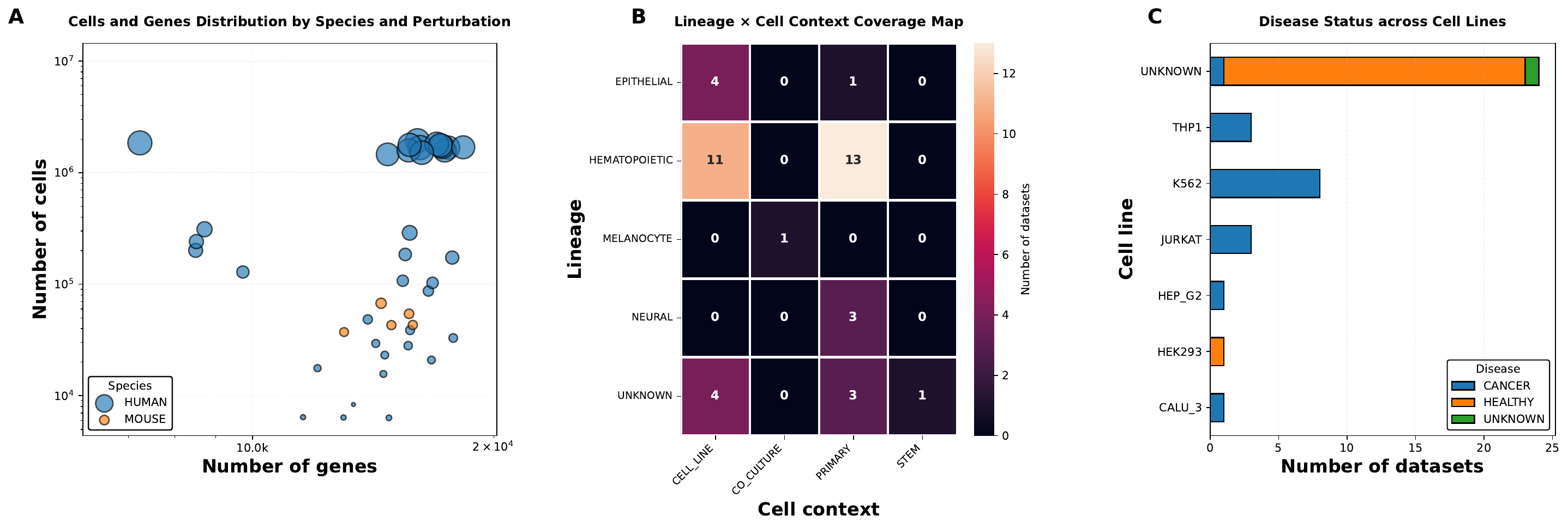}
  \caption{(A) shows the distribution of different datasets, the size of the circle represents the number of unique perturbations in this dataset. (B) The heatmap between lineage and cell context, numbers represent how many datasets are in each category combination. (C) The histogram for the number of datasets in each cell line. Colors indicate disease status, including cancer, healthy, and unknown. }
  \label{fig:dataapp}
\end{figure}
Perturbation information is stored separately from expression and metadata. Each cell is associated with a discrete perturbation token representing the orthology-aligned target gene or a dedicated \texttt{CONTROL} token for negative controls.
An explicit boolean flag identifies control cells, enabling efficient masking and stratification during training.

To support program-level analysis and aggregation, each cell is additionally assigned a deterministic program activity identifier derived from the dataset identity and perturbation token. This identifier is used solely for grouping and analysis and does not affect model inputs.

\subsection{Dataset Attributes and Provenance}

Each \texttt{zarr} dataset records global attributes including dataset name, number of cells, number of genes, creation timestamp, and a description of the storage format. These attributes enable transparent provenance tracking and facilitate automated dataset auditing.

\subsection{Consistency Verification and ``Zarr as Ground Truth''}

To guarantee correctness of the conversion process, we enforce a strict consistency check between the source \texttt{.h5ad} file and the generated \texttt{zarr} representation. Summary statistics include numbers of cells, genes, unique orthology tokens, unique perturbations, and unique values for each metadata field. These are computed independently from both representations and required to match exactly.

Only datasets that pass all consistency checks are admitted into the curated corpus used for pretraining and evaluation, with evaluation datasets held out from pretraining. The \texttt{zarr} representation is treated as the authoritative ground truth for all subsequent stages of data loading, tokenization, and model training.

\paragraph{Summary Construction.}
For each dataset, we construct a structured summary capturing dataset scale, vocabulary cardinalities, and metadata diversity. These summaries are aggregated across datasets to provide a global overview of corpus composition and to support downstream analysis and reporting.

\subsection{Gene Coverage In Pretraining}

\begin{figure}[t]
\centering

\begin{subfigure}[b]{0.48\linewidth}
    \centering
    \includegraphics[width=\linewidth]{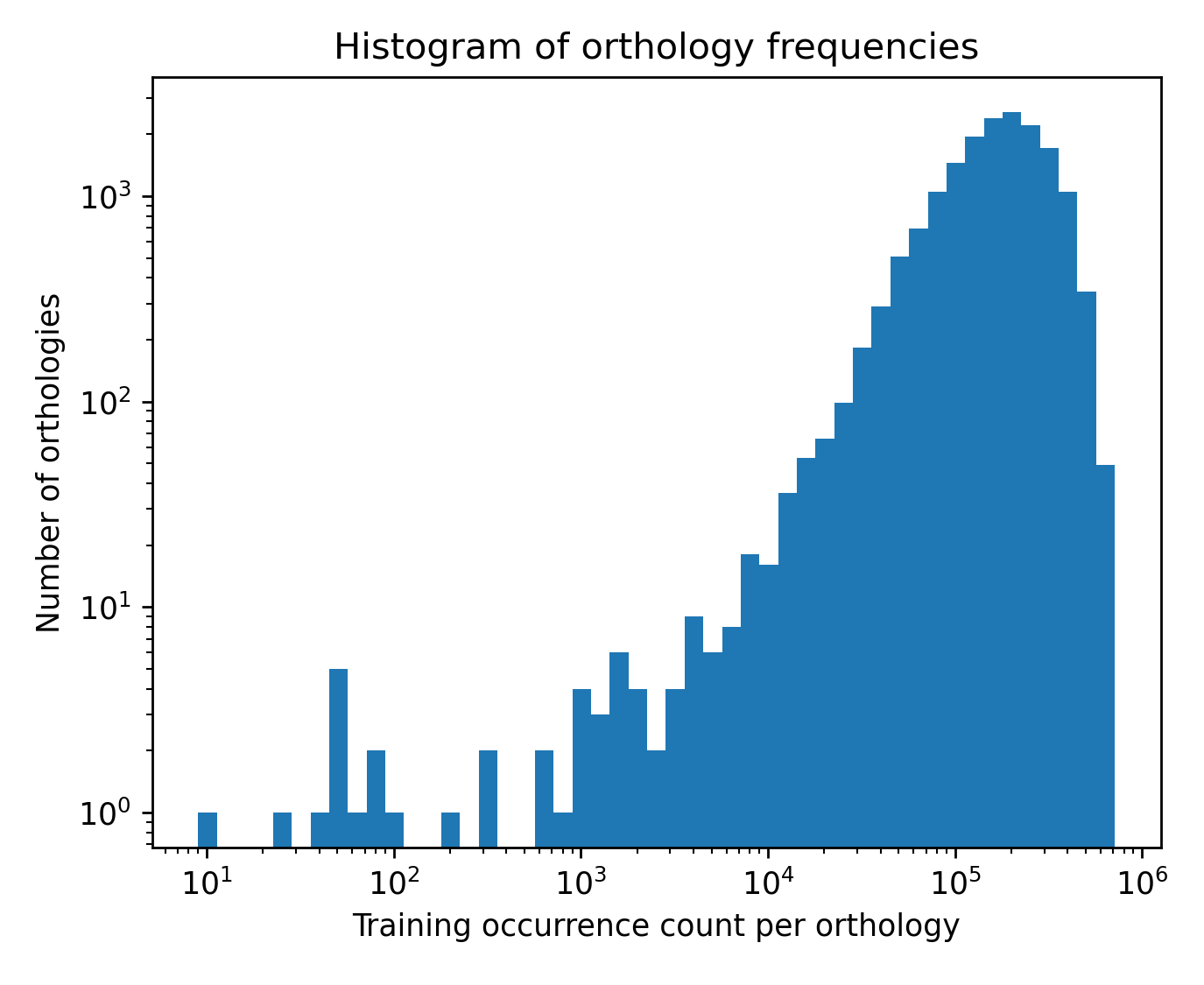}
    \caption{Stage 1 Gene Coverage}
\end{subfigure}
\hfill
\begin{subfigure}[b]{0.48\linewidth}
    \centering
    \includegraphics[width=\linewidth]{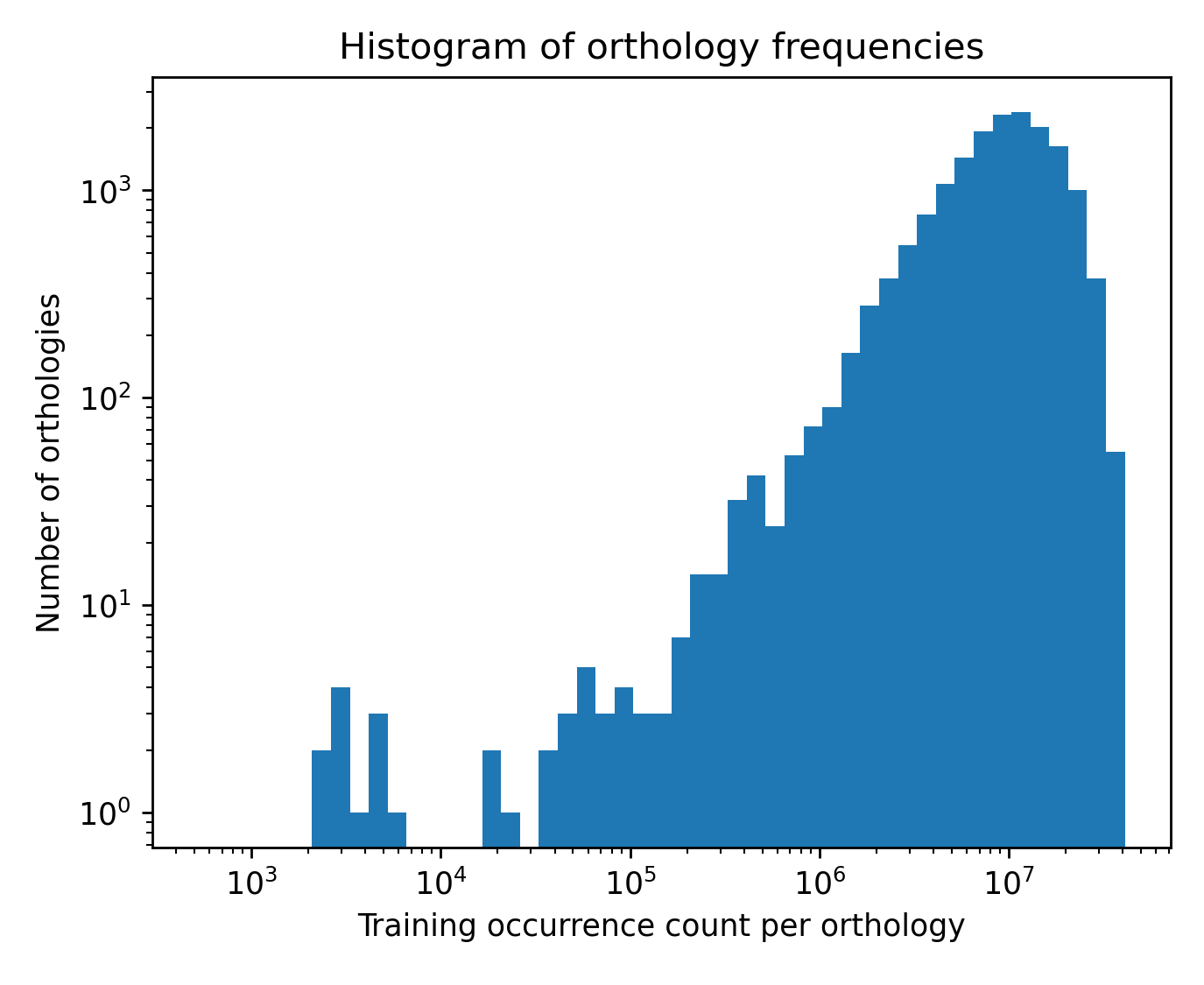}
    \caption{Stage 2 Gene Coverage}
\end{subfigure}

\caption{The gene coverage of training data with sparse tokenization}
\label{fig:two_images}
\end{figure}

Figure~\ref{fig:two_images} shows the distribution of gene occurrence frequencies during pretraining under sparse tokenization. 
In Stage~1, where training is performed on control cells, gene coverage is broadly distributed across the vocabulary, reflecting general transcriptional patterns. 
In Stage~2, perturbation-aware training shifts the distribution toward genes involved in perturbation responses, resulting in increased coverage of functionally relevant genes.

Despite the use of sparse tokenization, the overall gene vocabulary remains well covered across both stages. 
These results indicate that the pretraining process captures both general gene expression structure and perturbation-specific signals, supporting the ability of CytoWave to generalize across genes and experimental settings.

\begin{table*}[t]
\centering
\scriptsize
\caption{
Scale and perturbation coverage statistics for the full curated corpus before the dataset-held-out partitioning.
Datasets used for downstream evaluation are listed for completeness but are excluded from all pretraining stages.
}
\label{tab:pretrain_scale}
\setlength{\tabcolsep}{2pt}
\resizebox{\textwidth}{!}{
\begin{tabular}{lrrrrrr}
\toprule
&
&
&
&
\#Unique &
&
 \\
Dataset &
\#Cells &
\#Genes &
\#Orthologs &
Perturbations &
\#Control Cells &
\#Perturbed Cells \\
\midrule
ZhuMarson2025\_D1\_Rest & 1633989 & 16801 & 16801 & 11332 & 76543 & 1557446 \\
ZhuMarson2025\_D1\_Stim48h & 1534912 & 16815 & 16815 & 11403 & 73534 & 1461378 \\
ZhuMarson2025\_D1\_Stim8h & 1498205 & 16804 & 16804 & 11417 & 69632 & 1428573 \\
ZhuMarson2025\_D2\_Rest & 1851226 & 16805 & 16805 & 11307 & 83987 & 1767239 \\
ZhuMarson2025\_D2\_Stim48h & 1912736 & 16816 & 16816 & 11316 & 87747 & 1824989 \\
ZhuMarson2025\_D2\_Stim8h & 1948721 & 16808 & 16808 & 11469 & 88354 & 1860367 \\
ZhuMarson2025\_D3\_Rest & 1778418 & 16812 & 16812 & 11189 & 76823 & 1701595 \\
ZhuMarson2025\_D3\_Stim48h & 1762861 & 16811 & 16811 & 11081 & 77334 & 1685527 \\
ZhuMarson2025\_D3\_Stim8h & 1582510 & 16812 & 16812 & 11271 & 68441 & 1514069 \\
ZhuMarson2025\_D4\_Rest & 1630546 & 16813 & 16813 & 11639 & 76636 & 1553910 \\
ZhuMarson2025\_D4\_Stim48h & 1750133 & 16817 & 16817 & 11648 & 83474 & 1666659 \\
ZhuMarson2025\_D4\_Stim8h & 1637051 & 16817 & 16817 & 11688 & 77030 & 1560021 \\
DatlingerBock2017 & 5905 & 13987 & 13987 & 33 & 1320 & 4585 \\
DatlingerBock2021 & 39194 & 15467 & 15467 & 21 & 4497 & 34697 \\
DixitRegev2016\_K562\_13d & 15849 & 14143 & 14143 & 11 & 3491 & 12358 \\
DixitRegev2016\_K562\_7d & 28034 & 14587 & 14587 & 11 & 5381 & 22653 \\
DixitRegev2016\_K562\_HighMOI & 48763 & 14880 & 14880 & 11 & 8802 & 39961 \\
FrangiehIzar2021 & 191318 & 14694 & 14694 & 210 & 57605 & 133713 \\
GasperiniShendure2019 & 6680 & 13888 & 13888 & 1081 & 369 & 6311 \\
LaraAstiasoHuntly2023\_exvivo & 73564 & 15666 & 15666 & 42 & 12444 & 61120 \\
LaraAstiasoHuntly2023\_invivo & 53571 & 15465 & 15465 & 40 & 9096 & 44475 \\
LaraAstiasoHuntly2023\_leukemia & 45461 & 15471 & 15471 & 55 & 7717 & 37744 \\
LiangWang2023 & 35574 & 13614 & 13614 & 67 & 2459 & 33115 \\
NadigOConner2024\_HepG2 & 128850 & 9011 & 9011 & 2104 & 4976 & 123874 \\
NadigOConner2024\_Jurkat & 225309 & 8237 & 8237 & 2096 & 12013 & 213296 \\
NormanWeissman2019 & 103455 & 15550 & 15550 & 217 & 11855 & 91600 \\
PapalexiSatija2021\_ECCITE & 16708 & 12452 & 12452 & 22 & 2386 & 14322 \\
PapalexiSatija2021\_Arrayed & 6233 & 11629 & 11629 & 7 & 2009 & 4224 \\
ReplogleWeissman2022\_Essential & 295372 & 7913 & 7913 & 1960 & 10691 & 284681 \\
ReplogleWeissman2022\_GWPS & 1882585 & 7673 & 7673 & 9279 & 75328 & 1807257 \\
ReplogleWeissman2022\_RPE1 & 234223 & 8266 & 8266 & 2283 & 11485 & 222738 \\
SantinhaPlatt2023 & 41173 & 16638 & 16638 & 26 & 2520 & 38653 \\
ShifrutMarson2018 & 23107 & 13783 & 13776 & 19 & 3541 & 19566 \\
SunshineHein2023 & 83458 & 16373 & 16373 & 20321 & 19927 & 63531 \\
TianKampmann2019\_Neuron & 182790 & 16508 & 16508 & 41358 & 15580 & 167210 \\
TianKampmann2019\_iPSC & 275708 & 16211 & 16211 & 39162 & 10687 & 265021 \\
TianKampmann2021\_CRISPRa & 20400 & 16249 & 16249 & 96 & 434 & 19966 \\
TianKampmann2021\_CRISPRi & 30835 & 16304 & 16304 & 176 & 437 & 30398 \\
WesselsSatija2023 & 28117 & 15274 & 15263 & 170 & 424 & 27693 \\
XieHon2017 & 8371 & 14842 & 14842 & 52 & 7063 & 1308 \\
XuCao2023 & 97385 & 16755 & 16755 & 200 & 2758 & 94627 \\
\bottomrule
\end{tabular}
}
\end{table*}

\begin{table*}[t!]
\centering
\caption{
Biological and experimental context metadata across all curated datasets (with evaluation datasets held out from pretraining).
}
\label{tab:pretrain_context}
\setlength{\tabcolsep}{2pt}
\resizebox{\textwidth}{!}{
\begin{tabular}{llllll}
\toprule
Dataset &
Species &
Cell Context &
Cell Line &
Lineage &
Disease \\
\midrule
ZhuMarson2025\_D1\_Rest & Human & Primary & -- & Hematopoietic & Healthy \\
ZhuMarson2025\_D1\_Stim48h & Human & Primary & -- & Hematopoietic & Healthy \\
ZhuMarson2025\_D1\_Stim8h & Human & Primary & -- & Hematopoietic & Healthy \\
ZhuMarson2025\_D2\_Rest & Human & Primary & -- & Hematopoietic & Healthy \\
ZhuMarson2025\_D2\_Stim48h & Human & Primary & -- & Hematopoietic & Healthy \\
ZhuMarson2025\_D2\_Stim8h & Human & Primary & -- & Hematopoietic & Healthy \\
ZhuMarson2025\_D3\_Rest & Human & Primary & -- & Hematopoietic & Healthy \\
ZhuMarson2025\_D3\_Stim48h & Human & Primary & -- & Hematopoietic & Healthy \\
ZhuMarson2025\_D3\_Stim8h & Human & Primary & -- & Hematopoietic & Healthy \\
ZhuMarson2025\_D4\_Rest & Human & Primary & -- & Hematopoietic & Healthy \\
ZhuMarson2025\_D4\_Stim48h & Human & Primary & -- & Hematopoietic & Healthy \\
ZhuMarson2025\_D4\_Stim8h & Human & Primary & -- & Hematopoietic & Healthy \\
DatlingerBock2017 & Human & Cell line & Jurkat & Hematopoietic & Cancer \\
DatlingerBock2021 & Human & Cell line & Jurkat & Hematopoietic & Cancer \\
DixitRegev2016\_K562\_13d & Human & Cell line & K562 & Hematopoietic & Cancer \\
DixitRegev2016\_K562\_7d & Human & Cell line & K562 & Hematopoietic & Cancer \\
DixitRegev2016\_K562\_HighMOI & Human & Cell line & K562 & Hematopoietic & Cancer \\
FrangiehIzar2021 & Human & Co-culture & -- & Melanocyte & Unknown \\
GasperiniShendure2019 & Human & Cell line & K562 & Hematopoietic & Cancer \\
LaraAstiasoHuntly2023\_exvivo & Mouse & Primary & -- & Hematopoietic / Unknown & Healthy \\
LaraAstiasoHuntly2023\_invivo & Mouse & Primary & -- & Hematopoietic / Unknown & Healthy \\
LaraAstiasoHuntly2023\_leukemia & Mouse & Primary & -- & Hematopoietic / Unknown & Cancer \\
LiangWang2023 & Mouse & Cell line & -- & Epithelial & Healthy \\
NadigOConner2024\_HepG2 & Human & Cell line & HepG2 & Epithelial & Cancer \\
NadigOConner2024\_Jurkat & Human & Cell line & Jurkat & Hematopoietic & Cancer \\
NormanWeissman2019 & Human & Cell line & K562 & Hematopoietic & Cancer \\
PapalexiSatija2021\_ECCITE & Human & Cell line & THP1 & Hematopoietic & Cancer \\
PapalexiSatija2021\_Arrayed & Human & Cell line & THP1 & Hematopoietic & Cancer \\
ReplogleWeissman2022\_Essential & Human & Cell line & K562 & Unknown & Cancer \\
ReplogleWeissman2022\_GWPS & Human & Cell line & K562 & Unknown & Cancer \\
ReplogleWeissman2022\_RPE1 & Human & Cell line & -- & Epithelial & Healthy \\
SantinhaPlatt2023 & Mouse & Primary & -- & Epithelial & Healthy \\
ShifrutMarson2018 & Human & Primary & -- & Hematopoietic & Healthy \\
SunshineHein2023 & Human & Cell line & Calu-3 & Epithelial & Cancer \\
TianKampmann2019\_Neuron & Human & Primary & -- & Neural & Healthy \\
TianKampmann2019\_iPSC & Human & Stem & -- & Unknown & Healthy \\
TianKampmann2021\_CRISPRa & Human & Primary & -- & Neural & Healthy \\
TianKampmann2021\_CRISPRi & Human & Primary & -- & Neural & Healthy \\
WesselsSatija2023 & Human & Cell line & THP1 & Hematopoietic & Cancer \\
XieHon2017 & Human & Cell line & K562 & Unknown & Cancer \\
XuCao2023 & Human & Cell line & HEK293 & Unknown & Healthy \\
\bottomrule
\end{tabular}
}
\end{table*}

\begin{table*}[t!]
\centering
\caption{
Perturbation technologies and stimulation conditions across all curated datasets (with evaluation datasets held out from pretraining)
}
\label{tab:pretrain_perturb}
\setlength{\tabcolsep}{2pt}
\resizebox{\textwidth}{!}{
\begin{tabular}{llll}
\toprule
Dataset &
Perturbation Technology &
Stimulation &
Stimulation Time \\
\midrule
ZhuMarson2025\_D1\_Rest & CRISPRi & None & -- \\
ZhuMarson2025\_D1\_Stim48h & CRISPRi & TCR & 48h \\
ZhuMarson2025\_D1\_Stim8h & CRISPRi & TCR & 8h \\
ZhuMarson2025\_D2\_Rest & CRISPRi & None & -- \\
ZhuMarson2025\_D2\_Stim48h & CRISPRi & TCR & 48h \\
ZhuMarson2025\_D2\_Stim8h & CRISPRi & TCR & 8h \\
ZhuMarson2025\_D3\_Rest & CRISPRi & None & -- \\
ZhuMarson2025\_D3\_Stim48h & CRISPRi & TCR & 48h \\
ZhuMarson2025\_D3\_Stim8h & CRISPRi & TCR & 8h \\
ZhuMarson2025\_D4\_Rest & CRISPRi & None & -- \\
ZhuMarson2025\_D4\_Stim48h & CRISPRi & TCR & 48h \\
ZhuMarson2025\_D4\_Stim8h & CRISPRi & TCR & 8h \\
DatlingerBock2017 & CRISPR-KO & None / TCR & -- \\
DatlingerBock2021 & CRISPR-KO & None / TCR & -- \\
DixitRegev2016\_K562\_13d & CRISPR-KO & None & -- \\
DixitRegev2016\_K562\_7d & CRISPR-KO & None & -- \\
DixitRegev2016\_K562\_HighMOI & CRISPR-KO & None & -- \\
FrangiehIzar2021 & CRISPR-KO & None & -- \\
GasperiniShendure2019 & CRISPRi & None & -- \\
LaraAstiasoHuntly2023\_exvivo & CRISPR-KO & None & -- \\
LaraAstiasoHuntly2023\_invivo & CRISPR-KO & None & -- \\
LaraAstiasoHuntly2023\_leukemia & CRISPR-KO & None & -- \\
LiangWang2023 & CRISPR-KO & None & -- \\
NadigOConner2024\_HepG2 & CRISPRi & None & -- \\
NadigOConner2024\_Jurkat & CRISPRi & None & -- \\
NormanWeissman2019 & CRISPRa & None & -- \\
PapalexiSatija2021\_ECCITE & CRISPR-KO & None & -- \\
PapalexiSatija2021\_Arrayed & CRISPR-KO & None & -- \\
ReplogleWeissman2022\_Essential & CRISPRi & None & -- \\
ReplogleWeissman2022\_GWPS & CRISPRi & None & -- \\
ReplogleWeissman2022\_RPE1 & CRISPRi & None & -- \\
SantinhaPlatt2023 & CRISPR-KO & None & -- \\
ShifrutMarson2018 & CRISPR-KO & None / TCR & -- \\
SunshineHein2023 & CRISPRi & None & -- \\
TianKampmann2019\_Neuron & CRISPRi & None & -- \\
TianKampmann2019\_iPSC & CRISPRi & None & -- \\
TianKampmann2021\_CRISPRa & CRISPRa & None & -- \\
TianKampmann2021\_CRISPRi & CRISPRi & None & -- \\
WesselsSatija2023 & CRISPR-KD & None & -- \\
XieHon2017 & CRISPRi & None & -- \\
XuCao2023 & CRISPRi & None & -- \\
\bottomrule
\end{tabular}
}
\end{table*}

\begin{table*}[t]
\centering
\caption{
Statistics of evaluation datasets used for downstream perturbation prediction tasks.
}
\label{tab:eval_datasets}
\setlength{\tabcolsep}{2pt}
\resizebox{\textwidth}{!}{
\begin{tabular}{lccccc}
\toprule
Dataset &
Train &
Valid &
Test &
\#Unique Perturbations &
Mean Cells / Perturb \\
\midrule
DatlingerBock2017 &
1634 & 191 & 206 &
32 &
63.47 \\
DatlingerBock2017 (stim) &
2022 & 255 & 277 &
32 &
79.81 \\
GasperiniShendure2019 &
3982 & 397 & 747 &
286 &
17.92 \\
PapalexiSatija2021 &
3377 & 420 & 427 &
6 &
704.00 \\
SantinhaPlatt2023 &
30910 & 3854 & 3889 &
25 &
1546.12 \\
\bottomrule
\end{tabular}
}
\end{table*}

\section{Dataset Statistics and Corpus Composition}
\label{app:dataset_stats}

We summarize the composition and statistical properties of the full curated corpus collected for CytoWave in Tables~\ref{tab:pretrain_scale}--\ref{tab:pretrain_perturb}.  
A subset of these datasets is held out for downstream evaluation and excluded from pretraining.  
All tables describe the \emph{same set of datasets}, but present complementary views focusing on (i) dataset scale and perturbation coverage, (ii) biological and experimental context, and (iii) perturbation technologies and stimulation conditions, respectively.

\subsection{Scale and perturbation coverage.}
Table~\ref{tab:pretrain_scale} reports the scale of each dataset in terms of cell count, gene coverage, and perturbation diversity.
The corpus spans over six orders of magnitude in dataset size, ranging from small, early Perturb-seq studies with a few thousand cells to recent large-scale screens containing millions of single cells.
Across datasets, the number of orthology-aligned genes closely matches the number of retained genes, reflecting the strict gene mapping and filtering procedure described in Appendix~\ref{app:data}.
Importantly, the corpus covers a wide spectrum of perturbation regimes, from targeted perturbation panels with tens of genes to genome-scale screens involving tens of thousands of perturbations.
Control cells constitute a non-trivial fraction of most datasets, enabling robust estimation of baseline expression states and perturbation-associated deviations.

\subsection{Biological and experimental context diversity.}
Table~\ref{tab:pretrain_context} characterizes the biological and experimental context associated with each dataset through standardized metadata tokens.
The curated corpus integrates data from both human and mouse experiments, spanning primary cells, cell lines, and stem-cell-derived systems, and covering multiple lineages including hematopoietic, epithelial, neural, and melanocytic contexts.
Despite substantial heterogeneity across studies, the use of a fixed metadata schema ensures that each dataset is represented in a consistent tokenized form.
This design allows the model to condition on biological context explicitly while avoiding dataset-specific feature engineering.

\subsection{Perturbation technologies and stimulation conditions.}
Table~\ref{tab:pretrain_perturb} summarizes the perturbation modalities and stimulation settings present in the corpus.
The datasets collectively cover CRISPR knock-out (KO), knock-down (KD), CRISPR interference (CRISPRi), and CRISPR activation (CRISPRa) technologies.
Several datasets include multiple stimulation conditions (e.g., resting versus stimulated states) and time points, providing systematic variation in cellular state under identical perturbation targets.
Such a structure provides systematic variation in cellular state under identical perturbation targets, allowing CytoWave to learn perturbation responses under diverse contextual and temporal conditions during pretraining.

\subsection{Implications for foundation model pretraining.}
Taken together, these statistics highlight that the pretraining corpus is not merely large, but also structurally diverse along multiple axes: scale, perturbation coverage, biological context, and experimental condition.
By organizing all datasets under a unified orthology-aligned and schema-enforced representation, CytoWave is trained to learn perturbation responses that generalize across studies, species, and experimental designs, rather than overfitting to any single dataset or perturbation regime.

\subsection{Evaluation Dataset Selection}
\label{sec:eval_dataset_selection}

To comprehensively evaluate the generalization capability of CytoWave, we select a diverse set of downstream evaluation datasets that span a wide range of biological contexts, perturbation scales, and experimental designs (Table~\ref{tab:eval_datasets}). These datasets are strictly excluded from the pretraining corpus. 
Our selection criteria are guided by three principles: (i) coverage of heterogeneous biological conditions, (ii) diversity in perturbation complexity and data scale, and (iii) compatibility with standardized perturbation prediction tasks.

Specifically, the chosen datasets include both cell-line and primary-cell experiments, multiple CRISPR modalities (CRISPR-KO and CRISPRi), and varying perturbation densities, ranging from small-scale transcription factor screens (e.g., DatlingerBock2017) to large-scale enhancer and regulatory perturbation studies (e.g., GasperiniShendure2019).
Together, these datasets cover perturbation regimes with as few as single-digit cells per perturbation and as many as over one thousand cells per perturbation, enabling robust evaluation across low- and high-data regimes.

We further include datasets with substantially different perturbation cardinalities, from fewer than ten unique perturbations (PapalexiSatija2021) to several hundred perturbations (GasperiniShendure2019), in order to assess the stability of model performance under both narrow and broad perturbation vocabularies.
This diversity is critical for evaluating whether a model can generalize beyond densely sampled perturbations and avoid overfitting to dataset-specific perturbation distributions.

\section{Evaluation Metrics}
\label{app:metrics}
We evaluate perturbation prediction performance by computing expression changes relative to the control profile and measuring their Pearson correlation (PCC). Measuring changes relative to the control profile isolates perturbation-specific responses and removes baseline expression variability across cells and conditions. This provides a more faithful assessment of whether the model captures the direction and magnitude of perturbation responses.

\paragraph{Notation.}
For a test example $i$ with control profile $\mathbf{x}_i$, ground-truth perturbed profile $\mathbf{y}_i$, and model prediction $\hat{\mathbf{y}}_i$, we define the ground-truth and predicted expression changes as:
\begin{equation}
\boldsymbol{\delta}_i = \mathbf{y}_i - \mathbf{x}_i, \qquad
\hat{\boldsymbol{\delta}}_i = \hat{\mathbf{y}}_i - \mathbf{x}_i.
\end{equation}
where $\mathbf{x}_i$ is sampled from control cells under the same condition and metadata group as $\mathbf{y}_i$.

\paragraph{$\Delta$PCC (all genes).}
For each test example $i$, we compute the Pearson correlation between predicted and ground-truth changes across all $G$ genes:
\begin{equation}
\Delta\mathrm{PCC}_i = \mathrm{PCC}(\hat{\boldsymbol{\delta}}_i,\, \boldsymbol{\delta}_i)
= \frac{\sum_{g=1}^{G}(\hat{\delta}_{i,g} - \bar{\hat{\delta}}_i)(\delta_{i,g} - \bar{\delta}_i)}
{\sqrt{\sum_{g=1}^{G}(\hat{\delta}_{i,g} - \bar{\hat{\delta}}_i)^2}\,\sqrt{\sum_{g=1}^{G}(\delta_{i,g} - \bar{\delta}_i)^2}},
\end{equation}
where $\bar{\delta}_i = \frac{1}{G}\sum_{g}\delta_{i,g}$ and $\bar{\hat{\delta}}_i = \frac{1}{G}\sum_{g}\hat{\delta}_{i,g}$ are the mean changes. We report the average $\Delta\mathrm{PCC}$ over all test examples.

\paragraph{$\Delta$PCC$_{20}$ (top-20 differentially expressed genes).}
To assess performance on the most biologically relevant genes, we evaluate on the subset of genes exhibiting the largest expression changes. For each test example $i$, let
\begin{equation}
\mathcal{T}_i = \operatorname{top\text{-}20}_{g \in \mathcal{G}}\, |\delta_{i,g}|
\end{equation}
denote the set of 20 genes with the largest absolute ground-truth change. We then compute:
\begin{equation}
\Delta\mathrm{PCC}_{20,i} = \mathrm{PCC}\!\left(\{\hat{\delta}_{i,g}\}_{g \in \mathcal{T}_i},\, \{\delta_{i,g}\}_{g \in \mathcal{T}_i}\right).
\end{equation}
This metric emphasizes the model's ability to correctly predict the direction and magnitude of the strongest perturbation responses, which are typically of greater biological interest.

\paragraph{MAE (mean absolute error).}
To measure the magnitude of prediction errors, we compute the mean absolute error between predicted and ground-truth perturbed expression values over genes with measurable expression activity. 
For each test example $i$, let
\begin{equation}
\mathcal{S}_i = \{ g \mid x_{i,g} > 0 \ \text{or} \ y_{i,g} > 0 \}
\end{equation}
denote the set of genes with non-zero expression in either the reference control or perturbed profile. 
MAE is defined as:
\begin{equation}
\mathrm{MAE}_i =
\frac{1}{|\mathcal{S}_i|}
\sum_{g \in \mathcal{S}_i}
\left| \hat{y}_{i,g} - y_{i,g} \right|.
\end{equation}
We report the average MAE over all test examples. 
This formulation avoids domination of the metric by trivial background zeros in sparse single-cell expression matrices while quantifying absolute prediction error on genes with measurable expression activity.
\section{Additional Experimental Results}
\label{app:pathway}
\begin{figure*}[t!]
  \centering
  \includegraphics[width=\textwidth]{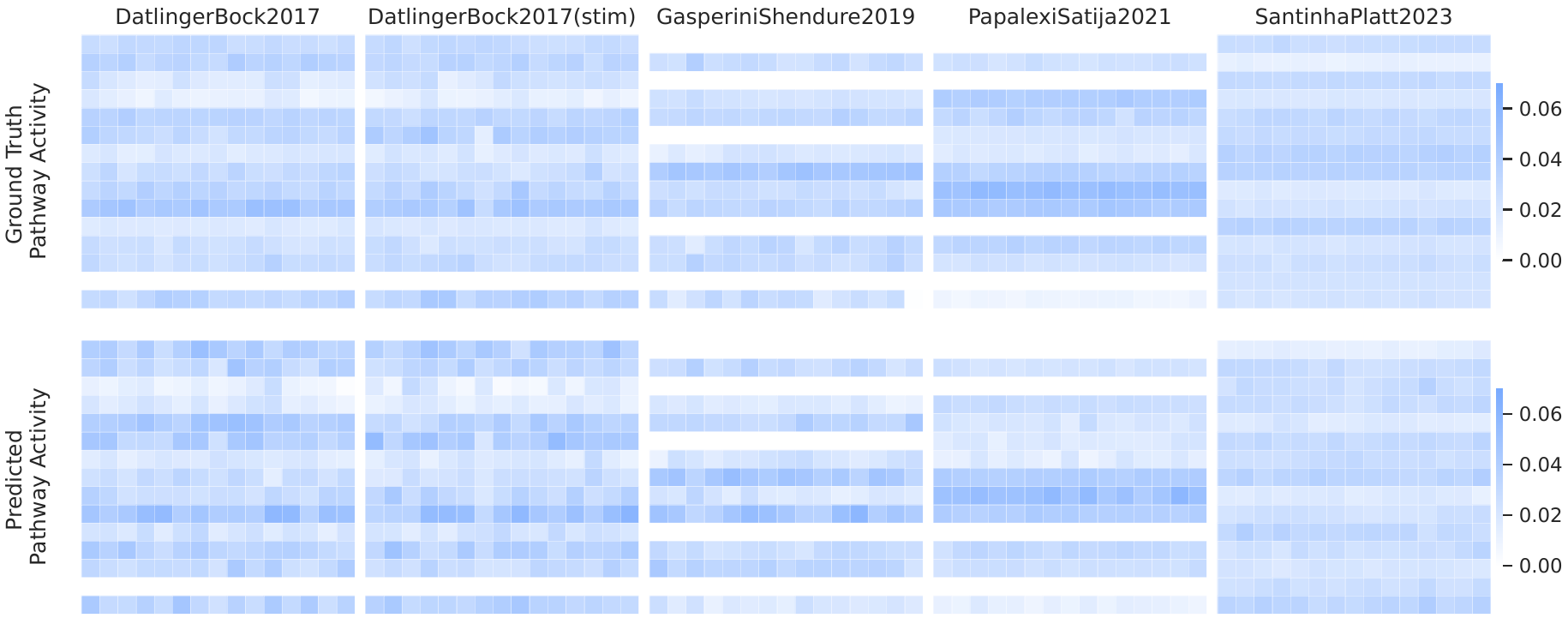}
  \caption{The top hallmark 50 absolute pathway activity of the perturbed ground truth and the predicted results, y-axis is hallmarks and x-axis is perturbation group in each dataset}
  \label{fig:absolute_pathway}
\end{figure*}

In Figure~\ref{fig:absolute_pathway}, we provide additional results for the pathway activity prediction. Figure~\ref{fig:absolute_pathway} visualizes absolute pathway activities for post-perturbation states, comparing ground-truth expression with model predictions. 
For each dataset, we first apply the trained CytoWave model to all test cells and obtain predicted post-perturbation gene expression profiles. 
To reduce cell-level noise and minimize redundant pathway computations, we aggregate gene expression by averaging across cells belonging to the same perturbation, yielding a perturbation-level bulk profile for both prediction and ground truth. 
Pathway activities are then computed once per perturbation using AUCell on orthology-mapped Hallmark gene sets. 
For visualization, we select the top 15 Hallmark pathways that exhibit the highest agreement between predicted and ground-truth activities across perturbations, and further select the top 15 perturbations with the strongest pathway-level activity in the ground truth. 
The resulting heatmaps display absolute pathway activities for the same set of pathways and perturbations, with the ground truth shown in the top row and the corresponding model predictions shown in the bottom row, using identical axes and color scales.

As shown in Figure~\ref{fig:absolute_pathway}, CytoWave accurately recovers absolute post-perturbation pathway activity patterns across diverse perturbations. 
The predicted heatmap closely matches the ground truth in both the relative ordering of perturbations within each pathway and the relative activation levels across pathways, indicating that the model preserves pathway-specific dynamic ranges rather than collapsing activities toward a global mean. 
Importantly, the strong visual alignment between the two rows demonstrates that CytoWave captures not only perturbation-dependent changes but also the correct post-perturbation activity landscape itself. 
These results complement the $\Delta$-activity analysis by showing that the model maintains faithful absolute pathway representations.

% ===========================
% Appendix: Implementation Details (CytoWave)
% ===========================
\section{Implementation Details}
\label{app:impl}

This section provides architectural and forward-pass details for CytoWave. We describe (i) the unified tokenization and embedding design, (ii) the bucketed joint transformer encoder for sparse scRNA inputs, and (iii) the stage-specific forward computations used in pretraining and finetuning.

\vspace{0.5em}
\paragraph{Unified vocabulary and role embeddings.}
All discrete symbols (genes, metadata tokens, and perturbation tokens) are mapped into a single unified vocabulary of size $|\mathcal{V}|$ and embedded by a shared table $E \in \mathbb{R}^{|\mathcal{V}|\times d}$ with model dimension $d$.
To distinguish token types, we add a learned \emph{role embedding} $R \in \mathbb{R}^{3\times d}$ with role IDs:
\texttt{ROLE\_GENE}=0, \texttt{ROLE\_META}=1, \texttt{ROLE\_PERTURB}=2.
Given token id $t$ and optional role id $r$, the embedding is
\begin{equation}
\mathrm{Embed}(t,r) \;=\; E[t] \;+\; \mathbbm{1}[r\ \text{provided}]\,R[r].
\end{equation}
Role embeddings are used consistently across all stages via \texttt{embed\_tokens()} (for meta/perturb supervision tokens) and \texttt{embed\_genes\_with\_expr()} (for sparse gene tokens with expression values).

\vspace{0.5em}
\paragraph{Expression FiLM for gene tokens.}
For a sparse gene token $g$ with observed log-normalized expression $x \in \mathbb{R}$, we modulate its base embedding $E[g]$ via an \emph{Expression FiLM} module:
\begin{equation}
(\gamma,\beta) \;=\; \mathrm{MLP}(x) \in \mathbb{R}^{2d},\qquad
\mathrm{FiLM}(E[g],x) \;=\; \gamma \odot E[g] + \beta,
\end{equation}
where $\mathrm{MLP}$ is a two-layer network \texttt{Linear(1,$h$)$\to$SiLU$\to$Dropout$\to$Linear($h$,2$d$)} and $h=\max(128,d)$ in our implementation. FiLM is applied \emph{only} to gene tokens; meta and perturb tokens use standard embeddings without FiLM. When role IDs are provided, role embeddings are added after FiLM:
\begin{equation}
\mathrm{EmbedGene}(g,x,r) \;=\; \mathrm{FiLM}(E[g],x) \;+\; \mathbbm{1}[r]\,R[r].
\end{equation}

\vspace{0.5em}
\paragraph{CSR input representation and joint attention.}
Each cell is represented as (i) sparse gene tokens in CSR form \texttt{(gene\_token\_ids, gene\_expr, gene\_batch\_ptr)} and (ii) a fixed-length sequence of meta tokens \texttt{meta\_token\_ids} (shape $B\times S$ for batch size $B$).
We first embed all nonzero gene entries with FiLM, producing a flat tensor of size $\mathrm{nnz}\times d$, then slice per cell using \texttt{gene\_batch\_ptr}.
For each cell $i$, we construct a variable-length sequence by concatenating meta and gene tokens:
\begin{equation}
\mathbf{s}_i \;=\; [\,\mathbf{m}_{i,1:S}\ ;\ \mathbf{g}_{i,1:L_i}\,] \in \mathbb{R}^{(S+L_i)\times d},
\end{equation}
and encode it with a shared transformer set encoder (\S\ref{app:impl:bucket}). The encoder output is mean-pooled over non-padding tokens to obtain the cell embedding $z \in \mathbb{R}^{B\times d}$.
We also compute a pooled meta state by simple mean over meta tokens:
\begin{equation}
\mathrm{meta\_state}_i \;=\; \frac{1}{S}\sum_{j=1}^{S}\mathbf{m}_{i,j}.
\end{equation}
The main encoding routine \texttt{encode\_csr()} returns $(z, \mathrm{meta\_state})$.

\vspace{0.5em}
\paragraph{Bucketed variable-length transformer encoding.}
\label{app:impl:bucket}
To reduce padding overhead for highly variable $L_i$, we implement length-based bucketing: sequences are sorted by length, partitioned into buckets of size \texttt{tf\_bucket\_size}, padded within each bucket, encoded, and then restored to the original order. Concretely, for a bucket of $b$ sequences with max length $L_{\max}$, we build
\texttt{x\_padded} $\in \mathbb{R}^{b\times L_{\max}\times d}$ and a boolean \texttt{key\_padding\_mask} $\in \{0,1\}^{b\times L_{\max}}$ (\texttt{True} denotes padding positions), run \texttt{nn.TransformerEncoder} with \texttt{src\_key\_padding\_mask}, and mean-pool over valid tokens. When available, we enable PyTorch 2.x SDPA backends (FlashAttention and memory-efficient attention) in a best-effort manner.

\vspace{0.5em}
\paragraph{Decoder heads.}
We decode gene-level expression for a set of target gene tokens using a shared \texttt{ExprDecoder}:
\texttt{LayerNorm$\to$Linear($d$,$d$)$\to$SiLU$\to$Linear($d$,1)} followed by \texttt{softplus}, with the final linear layer initialized to zeros (both weight and bias) for stable early training.

\vspace{0.5em}
\paragraph{Perturbation set encoding.}
A (possibly multi-gene) perturbation set per cell is provided in CSR form \texttt{(perturb\_token\_id, perturb\_batch\_ptr)}. We embed all perturbation tokens using \texttt{embed\_tokens()} (optionally with role IDs) and perform mean-pooling within each cell using scatter-add:
\begin{equation}
e_u^{(i)} \;=\; \frac{1}{|\mathcal{P}_i|}\sum_{p \in \mathcal{P}_i} \mathrm{Embed}(p, \mathrm{ROLE\_PERTURB}),
\end{equation}
returning $e_u \in \mathbb{R}^{B\times d}$. If no perturbation tokens are present, we return zeros.

\vspace{0.5em}
\paragraph{Stage 1: masked gene recovery on control cells.}
Given encoded control embedding $z_0 \in \mathbb{R}^{B\times d}$ and a masked set of gene tokens (CSR), we predict masked gene expression by adding $z_0$ as a global residual to each masked token embedding:
\begin{equation}
\hat{x}_{i,k} \;=\; \mathrm{Dec}\!\left(\mathrm{Embed}(g_{i,k},\mathrm{ROLE\_GENE}) + z_{0}^{(i)}\right),
\end{equation}
where $\mathrm{Dec}$ is \texttt{ExprDecoder}. This is implemented by iterating over cells and concatenating predictions for all masked targets.

\vspace{0.5em}
\paragraph{Stage 2: latent perturbation operator.}
We encode the control cell to obtain $z_x$ and encode the perturbation set to obtain $e_u$. The operator predicts an additive latent state change using an MLP:
\begin{equation}
\Delta z \;=\; f_{\mathrm{op}}([z_x;e_u]), \qquad z' \;=\; z_x + \Delta z,
\end{equation}
where $f_{\mathrm{op}}$ is \texttt{MLP(LN$\to$Linear$\to$SiLU$\to$Dropout$\to$Linear)} with input dimension $2d$ and output dimension $d$.
We then decode target gene expression using the same token+cell residual mechanism as Stage 1. For efficiency, decoding over all supervised target tokens is vectorized by broadcasting $z'$ to each target gene token via \texttt{repeat\_interleave} on the CSR pointers, avoiding per-cell loops.

\vspace{0.5em}

We encode the target (post-perturbation) cell representation $z_y$ using \texttt{encode\_csr()} and form a latent difference
\begin{equation}
\delta \;=\; z_y - z_x.
\end{equation}
We also build a context vector by summing pooled meta state and an embedding of the \emph{observed} perturbation subset $e_{\mathrm{obs}}$ (when provided, via masked perturbation CSR; otherwise zeros). The identifier predicts a perturbation embedding
\begin{equation}
\hat{e} \;=\; f_{\mathrm{id}}([\delta;\ \mathrm{meta\_state}+e_{\mathrm{obs}}]),
\end{equation}
where $f_{\mathrm{id}}$ is an MLP analogous to $f_{\mathrm{op}}$.
Supervision uses a target perturbation embedding $e_{\mathrm{tgt}}$ computed from \texttt{target\_perturb\_*} if available, otherwise falling back to the full perturbation set \texttt{perturb\_*}. In our implementation, Stage 2 can optionally return both the operator outputs and $(\hat{e}, e_{\mathrm{tgt}})$ within a single forward pass to avoid redundant encoding of $z_x$.

\vspace{0.5em}
\paragraph{Stage 4: finetuning forward.}
For downstream prediction, we use the same operator and decoding as Stage 2, but decode \emph{all} target gene tokens. 

\vspace{0.5em}
\paragraph{Shapes and batch pointers.}
Throughout, sparse gene tokens are represented as flat arrays of length $\mathrm{nnz}$ with a batch pointer \texttt{ptr} of length $B{+}1$, where \texttt{ptr[i]:ptr[i+1]} indexes tokens belonging to cell $i$. We use this convention for control genes, target genes, masked gene supervision, and perturbation sets.

\paragraph{Baseline implementation.}
All baselines are trained and evaluated using the same train/validation/test splits. For pretrained baselines, we follow their released finetuning protocols when available and adapt them only on the held-out training split. Hyperparameters are selected using validation performance, and test results are reported after model selection.

\paragraph{Pretraining Setup}
CytoWave is pretrained by a two-stage self-supervised strategy. In Stage 1, the model learns perturbation-agnostic representations from control cells through masked gene expression recovery with a masking ratio of 15\%. The encoder is implemented as an 8-layer Transformer with 8 attention heads and model dimension $d=512$. CytoWave is pretrained 20 epochs using the AdamW optimizer with a learning rate of 2e-4, weight decay of 0.01, and gradient clipping at 1.0. A cosine learning rate schedule with linear warmup is applied. In Stage 2, CytoWave conducts perturbation-aware representation learning on top of the pretrained encoder from stage 1. We use a 30\% masking ratio in perturbation response expression mask and recovery. The designed masking first includes the top 300 gene expressions with the largest changes after perturbations and then randomly samples from the rest. The model predicts post-perturbation expression values for this selected gene set, which are later evaluated through expression changes relative to the control profile while simultaneously learning to recover masked perturbation targets. The total training loss is defined as the sum of the expression regression loss and the perturbation recovery loss, with a weighting factor of 5.0 applied to the latter. Stage 2 is trained for 10 epochs with a batch size of 24 and gradient accumulation of 4, using the same optimizer and learning rate as Stage 1. The InfoNCE denominator is computed within each mini-batch and does not gather negatives across gradient accumulation steps. Mixed-precision training with bfloat16 is enabled to improve training efficiency. All pretraining experiments were conducted on a cluster with 8 NVIDIA H100 GPUs using distributed data parallel training. The complete two-stage pretraining pipeline required approximately two weeks of training time.

\section{Limitation}
\label{app:limitation}
This study suggests that genetic perturbation data provides a qualitatively different and richer training signal than unperturbed single-cell atlases, as perturbations directly encode how cellular states respond under controlled perturbations. Representing perturbation responses in a shared latent space provides a structured approach to distinguish between intrinsic cell-state variation and perturbation-dependent responses. A key implication of this design is improved generalization in settings where direct supervision is scarce. By learning perturbation-dependent responses as latent state changes rather than dataset-specific conditional regressions, the model can reuse and compose learned response patterns across experimental conditions.

Several limitations should be noted. Perturbation coverage remains uneven across genes and cell types, which may limit the model’s ability to capture gene-specific responses consistently across different contexts. Because CytoWave uses an orthology-aligned vocabulary, genes without valid orthology assignments are excluded, which may limit modeling of species-specific regulatory programs. In addition, while the learned representations reflect perturbation-associated patterns at the gene and pathway levels, they do not enforce explicit mechanistic constraints. Understanding how these representations relate to known regulatory or causal structures remains an open question for future investigation.

Future work may extend CytoWave along several directions, including expanding perturbation coverage, incorporating multi-omic measurements, and exploring richer perturbation operators that capture temporal and nonlinear responses. We believe that such developments are essential for building foundation models that not only predict perturbation outcomes, but also support mechanistic understanding and hypothesis generation in complex biological systems.

\section{Broader Impact}

CytoWave is designed for modeling single-cell genetic perturbation responses and may support applications in functional genomics, disease modeling, and therapeutic discovery. By improving the ability to predict perturbation-associated transcriptional responses, such models may help prioritize experimental hypotheses and reduce experimental cost in biomedical research.

We do not identify direct negative societal impacts associated with the proposed method beyond those generally related to machine learning models used in biological research. Incorrect perturbation predictions may nevertheless lead to misleading biological hypotheses if used without experimental validation. CytoWave is intended as a research tool and does not provide clinical or medical decision-making recommendations.

\section{Dataset Licenses and Terms of Use}

All datasets were obtained from publicly available academic resources and remain subject to their original licenses and terms of use. Where available, datasets follow licenses and academic usage terms provided by their original repositories or publications, including common academic research licenses such as CC-BY and related non-commercial usage terms.
% \section{Technical appendices and supplementary material}
% Technical appendices with additional results, figures, graphs, and proofs may be submitted with the paper submission before the full submission deadline (see above). You can upload a ZIP file for videos or code, but do not upload a separate PDF file for the appendix. There is no page limit for the technical appendices. 

% Note: Think of the appendix as ``optional reading'' for reviewers. The paper must be able to stand alone without the appendix; for example, adding critical experiments that support the main claims to an appendix is inappropriate. 

% %%%%%%%%%%%%%%%%%%%%%%%%%%%%%%%%%%%%%%%%%%%%%%%%%%%%%%%%%%%%

\newpage

\end{document}